\pdfoutput=1
\documentclass[10pt, logo, copyright]{nvidiatechreport}
\usepackage[square,numbers]{natbib}
\usepackage{amsmath,amsfonts,bm}

\def\eqref#1{equation~\ref{#1}}

\def\1{\bm{1}}

\DeclareMathAlphabet{\mathsfit}{\encodingdefault}{\sfdefault}{m}{sl}
\SetMathAlphabet{\mathsfit}{bold}{\encodingdefault}{\sfdefault}{bx}{n}

\usepackage{url}
\usepackage{graphicx}
\usepackage{tikz}
\usepackage{booktabs,tabularx}
\usepackage{multirow}
\usepackage{wrapfig}
\usepackage{placeins}
\usepackage{amsmath,amssymb}
\usepackage{xcolor}
\usepackage{booktabs}
\usepackage{listings}
\usepackage{etoc}

\usetikzlibrary{arrows.meta,backgrounds,calc,fit,positioning}

\title{TAMPAgent: Agentic Belief-Space Planning for Mobile Manipulation}

\author[1,2]{Utkarsh A. Mishra}
\author[1]{Alperen Degirmenci}
\author[2]{Yongxin Chen}
\author[1,2]{Danfei Xu}
\author[1,3]{Tucker Hermans}
\author[1]{Caelan Garrett}
\affil[1]{NVIDIA Research}
\affil[2]{Georgia Institute of Technology}
\affil[3]{University of Utah}

\paperurl{https://umishra.me/TAMPAgent/}

\makeatletter
\gdef\theabstract{{\def\abstract{}\def\endabstract{}
  \textbf{Abstract:} \begin{abstract}
Open-world robotic mobile manipulation requires acting in physical environments with only partial information. While recent Large Vision-Language Model (LVLM) agents show strong open-world reasoning capabilities, existing architectures that rely on direct tool calling struggle with geometric uncertainty and thus are limited to fully-observable tabletop environments. In contrast, Task and Motion Planning (TAMP) systems can solve long-horizon tasks under partial observability but incur high computational costs and require closed worlds. We propose \textbf{TAMPAgent}, an agentic planning framework that unifies the reasoning of LVLMs with the structured constraint satisfaction of TAMP systems. We endow an LVLM agent with a belief-space planning model operating over an extensible API of perception, planning, and control tools. The agent formulates a partial task plan composed of parameterized observation and execution actions, which is subsequently completed using classical search algorithms. After extracting the plan's geometric constraints, the agent leverages its tool API to bind valid parameter values, satisfying feasibility requirements and producing an executable motion plan. Upon execution, the agent continuously updates its belief history and repeats this closed-loop process by interleaving observation and manipulation actions. Evaluated across complex household mobile manipulation tasks on Rainbow Y1 and Unitree G1 platforms, TAMPAgent significantly outperforms direct tool-calling agentic baselines, demonstrating that integrating TAMP constraint satisfaction into agentic planning loops is essential for unstructured, partially observable environments.

\end{abstract}}%
  \par\noindent
  \textbf{Keywords:} Task and Motion Planning, Agentic Robotics, Mobile Manipulation
  \par\noindent
  \textbf{Website:} \href{https://umishra.me/TAMPAgent/}{https://umishra.me/TAMPAgent/}}
\makeatother

\usepackage{listings}

\usepackage[most]{tcolorbox}
\newtcblisting{stepbox}[2]{listing only, title={#2}, enhanced, breakable=false,
  colback=gray!8, colframe=gray!55, colbacktitle=gray!22, coltitle=black,
  fonttitle=\scriptsize\bfseries, boxrule=0.4pt, titlerule=0.4pt, arc=2pt,
  left=5pt, right=5pt, top=4pt, bottom=4pt, toptitle=2.5pt, bottomtitle=2.5pt, before skip=8pt, after skip=8pt,
  listing options={style=#1, escapeinside={(*@}{@*)}, frame=none, backgroundcolor={}, xleftmargin=0pt, xrightmargin=0pt, aboveskip=0pt, belowskip=0pt}}
\lstdefinestyle{lean}{
  basicstyle=\ttfamily\scriptsize,
  alsoletter={},
  emph={def,variable,theorem,example,by,exact,obtain,Prop,Type},
  emphstyle=\bfseries,
  mathescape=false,
  escapeinside={},
  literate=
    {∃}{{$\exists$}}1 {∀}{{$\forall$}}1 {∧}{{$\wedge$}}1
    {→}{{$\rightarrow$}}1 {⟨}{{$\langle$}}1 {⟩}{{$\rangle$}}1
    {τ}{{$\tau$}}1 {Φ}{{$\Phi$}}1
    {₀}{{$_0$}}1 {₁}{{$_1$}}1 {₂}{{$_2$}}1 {₃}{{$_3$}}1
}

\definecolor{kw}{RGB}{0,90,90}
\definecolor{cm}{RGB}{110,110,110}

\lstdefinestyle{pddl}{
  basicstyle=\small\ttfamily,
  columns=fullflexible,
  breaklines=true,
  frame=single, framerule=0.3pt,
  xleftmargin=4pt, xrightmargin=4pt,
  keywordstyle=\color{kw}\bfseries,
  morekeywords={define,domain,problem,and,not,or},
  commentstyle=\color{cm}\itshape,
  morecomment=[l]{;;},
  showstringspaces=false,
  mathescape=false,
  escapeinside={}
}
\lstdefinestyle{trace}{
  basicstyle=\ttfamily\scriptsize, columns=fixed, basewidth=0.5em, keepspaces=true,
  emph={}, mathescape=false, escapeinside={}, showstringspaces=false, breaklines=false,
  backgroundcolor=\color{gray!10}, frame=single, framerule=0.3pt, xleftmargin=1em, xrightmargin=1em
}
\lstdefinestyle{rec}{
  basicstyle=\scriptsize\ttfamily,
  emph={},
  keepspaces=true,
  columns=fullflexible,
  breaklines=true,
  frame=single, framerule=0.3pt,
  xleftmargin=4pt, xrightmargin=4pt,
  showstringspaces=false,
  mathescape=false,
  escapeinside={}
}

\newcommand{\pddl}[1]{{\texttt{#1}}} %

\newif\ifcomments
\commentstrue
\ifcomments
    \newcommand{\caelan}[1]{\textcolor{blue}{(CG: #1)}}
    \newcommand{\trh}[1]{\textcolor{orange}{(TH: #1)}}
    \newcommand{\utkarsh}[1]{\textcolor{cyan}{(UM: #1)}}
    \newcommand{\danfei}[1]{\textcolor{brown}{(DX: #1)}}
    \newcommand{\yongxin}[1]{\textcolor{red}{(YC: #1)}}
    \newcommand{\slide}[1]{\textcolor{gray}{\footnotesize[slide~#1]}}
    \newcommand{\todoitem}[1]{\textcolor{red}{\textbf{[TODO: #1]}}}
\else
    \newcommand{\caelan}[1]{}
    \newcommand{\trh}[1]{}
    \newcommand{\utkarsh}[1]{}
    \newcommand{\danfei}[1]{}
    \newcommand{\yongxin}[1]{}
    \newcommand{\slide}[1]{}
    \newcommand{\todoitem}[1]{}
\fi

\begin{document}
\def\ArxivMainText{}
\maketitle
\etocdepthtag.toc{mtmain}

\section{Introduction}
\label{sec:introduction}

Robots in homes must solve long-horizon manipulation tasks in environments they have never mapped, where target objects may lie in other rooms, behind clutter, or inside closed containers. Such tasks couple deciding what to find out with deciding how to act: the robot must choose where to look, what to move out of the way, and how to grasp an object so that it can later be placed, often before it has seen where the object will go. Large Vision-Language Model (LVLM) agents~\citep{anthropic2026opus55,google2026gemini4argon} have recently become a popular way to control robots due to their simplicity.
Given perception, grasping, and motion tools, an agent can write the policy directly as a program that orchestrates tool calls~\citep{liang2023codeaspolicies,huang2023voxposer,chen2026gap,galanti2026addressing}, bringing commonsense priors about where objects are likely to be and the semantic flexibility to compose new behaviors. However, these programs, whether open-loop scripts~\citep{liang2023codeaspolicies} or fixed execution graphs~\citep{chen2026gap}, call tools without explicitly representing the geometric or probabilistic constraints each action must satisfy and thus often fail when deployed zero-shot on a new task.
 
Task and Motion Planning (TAMP)~\citep{garrett2021integratedtamp} 
explicitly reasons over parameterized actions whose preconditions constrain grasps, poses, and trajectories, so that a plan can be verified before it is executed.
The majority of TAMP systems, however, make two assumptions that break in open-world settings: (1) hand-authored problems and (2) fully observed scenes. 
Recently, VLM-TAMP systems relax (1) by letting agents translate instructions and images into the goals, objects, and constraints of a planning problem~\citep{yang2024vlmtamp,kumar2024owltamp}, but a classical solver still orchestrates the search, confining the agent to a rigid sequence of operations in which it is often queried only once before execution~\citep{shen2026tiptop} in predominantly observable environments. Belief-space TAMP~\citep{kaelbling2013belieftamp,garrett2020onlinereplanning,curtis2024tampura} relaxes (2) by treating beliefs, distributions over states, as action parameters and observation as a type of action, 
but it represents beliefs as rigid parametric or non-parametric distributions that are propagated with fixed and complex sampling and inference procedures.
In contrast, humans use simpler concepts rooted in language to represent uncertainty, such as ``the mug is probably in the upper cabinet'', and are able to induce the same decision about where to look first as a full distribution would.

\begin{figure}[t!]
\centering
\includegraphics[width=\linewidth]{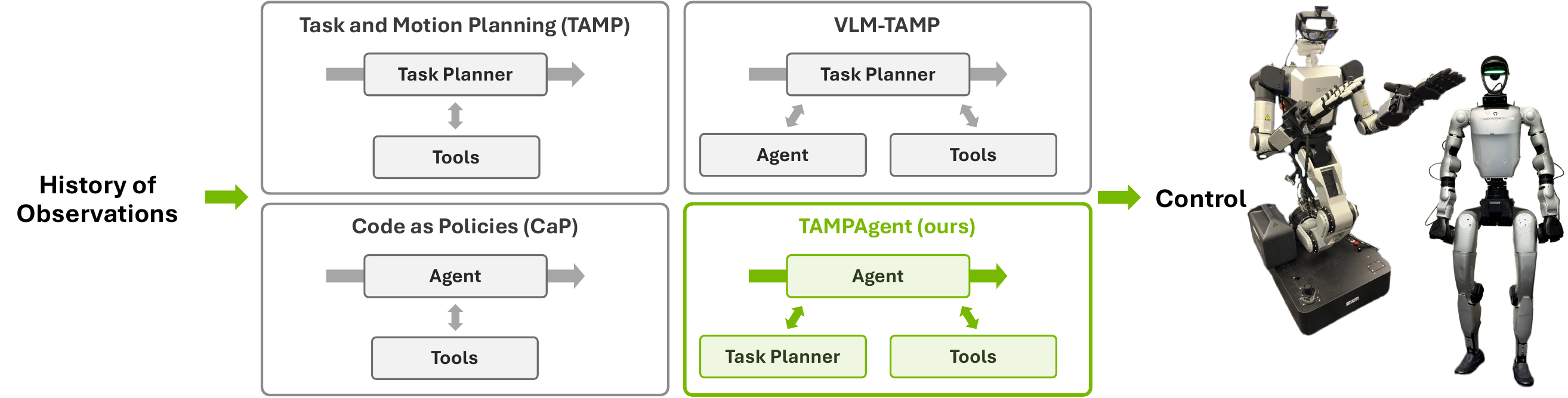}
\caption{\textbf{Integrating agents with planning.} We compare four systems that maps a history of observations to whole body control for long-horizon mobile-manipulation tasks: (a) \textbf{TAMP}~\citep{garrett2020pddlstream,garrett2021integratedtamp} solves hand-authored problems with a task planner that orchestrates tools. (b) \textbf{VLM-TAMP}~\citep{kumar2024owltamp,shen2026tiptop} uses the agent to write the problem from open-world understanding but keeps the planning pipeline fixed. (c) \textbf{CaP}~\citep{liang2023codeaspolicies,lu2026aspire,chen2026gap,galanti2026addressing} lets the agent write programs that orchestrate tools directly to find solutions. (d) \textbf{TAMPAgent} (ours) lets the agent orchestrate both the task planner and the tools to find certified solutions that satisfy the planner derived constraints.
}
\label{fig:teaser}
\end{figure}

We propose \emph{belief-space agentic-TAMP} (\textbf{TAMPAgent}), which 
integrates the strengths of LVLMs and TAMP
by having an agent orchestrate both belief-space task planning and tool usage (\autoref{fig:teaser}). 
The agent acts as the belief-space policy with full authority over tool use, while its output keeps the contract of classical TAMP: parameterized actions whose values must satisfy constraints declared by the planner. Because the agent outputs plans, not programs, each action's effects induce a subsequent belief state, so the agent can reason directly about what will be true and what it will know in the future, such as a placement that becomes certifiable only once the plate has been observed. Beliefs are hybrid: each plan variable, from an object's location to a grasp or a placement, takes either a language hypothesis written by the agent or a continuous value produced by a perception, grasping, or motion tool. Over language beliefs, the agent handles constraints and costs implicitly, avoiding the explicit integration over beliefs that belief-space TAMP requires, and tool-certified continuous values are required before any motor execution. We evaluate TAMPAgent on simulated household tasks with a Rainbow Y1~(RBY1) and a Unitree G1~(G1) robot. On the RBY1 kitchen tasks, it reaches 77.2\% absolute success, against 57.2\% for CaP~\cite{liang2023codeaspolicies} and 61.1\% for Graph as Policy (GaP)~\cite{chen2026gap}, while using fewer uncached input and output tokens per minute of simulated time than both. We also demonstrate our method on three categories of real-world mobile manipulation using RBY1 and on clearing a blocker with a real G1. \textbf{Our contributions are:} (i) the belief-space TAMPAgent formulation, in which an agent serves as the belief-space policy while a planner compiles its decisions into verifiable constraints and (ii) hybrid beliefs that combine language and continuous values for every plan variable.

\section{Preliminaries: Belief-Space Task and Motion Planning}
\label{sec:preliminaries}

We first review classical, {\it i.e.}, deterministic observable, Task and Motion Planning (TAMP)~\cite{garrett2021integratedtamp} and then belief-space TAMP~\cite{kaelbling2013belieftamp}.
In \autoref{sec:method}, we will marry these ideas with modern LVLM agents.

\begin{wrapfigure}{r}{0.45\linewidth}
\centering
\vspace{-4em}
\includegraphics[width=\linewidth]{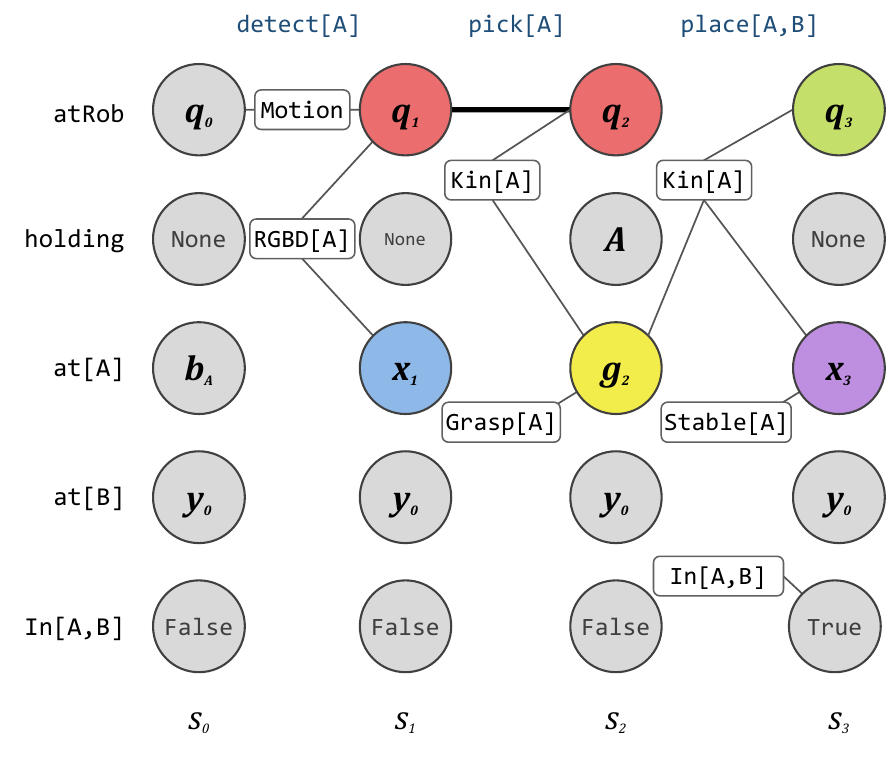}
\caption{\textbf{Plan skeleton as a factor graph.} The skeleton $\vec{a} = \langle \pddl{detect}[A], \pddl{pick}[A], \pddl{place}[A,B] \rangle$ unrolled over states $s_0, \dots, s_3$ (columns) becomes factor graph with: (i) nodes $V$ as the planning variables (rows), such as the configurations $q_1, q_2, q_3$, the apple's pose, the grasp $g_2$, and the placement $x_3$ and (ii) factors $F$ as the constraints between them, such as \pddl{RGBD}, \pddl{Grasp}, \pddl{Kin}, \pddl{Motion}, and \pddl{Stable} visualized in boxes. The apple's pose starts as a belief $b_A$ and becomes the observed pose $x_1$ through \pddl{detect} action.
}
\vspace{-1em}
\label{fig:factor-graph}
\end{wrapfigure}

\paragraph{Task and motion planning.}
In TAMP, the world is described by state variables, such as the robot configuration $\pddl{atRob}$, the held object $\pddl{holding}$, the pose $\pddl{at}[o]$ of each object $o$, and relations such as $\pddl{In}[o,o']$, which is \texttt{True} when $o$ is inside $o'$. The state $s_t$ at time $t$ assigns a value to each variable.
TAMP actions $a$ have discrete arguments (\textit{what behavior, which object}) and continuous parameters (\textit{how}). Each action has preconditions (\pddl{pre:}) and effects (\pddl{eff:}) on the state variables: \pddl{pick}$[o]$, for instance, requires $\pddl{holding} = \texttt{None}$ and sets $\pddl{holding} = o$. It also specifies \emph{constraints} (\pddl{con:}) on its continuous parameters: a stable grasp (\pddl{Grasp}), reachable from a robot configuration (\pddl{Kin}) through a collision-free path (\pddl{Motion}). Likewise, \pddl{place}$[o,o']$ requires $\pddl{holding} = o$ and a stable pose (\pddl{Stable}) on its support $o'$. 
Consider an example task where a robot must place an apple $A$ in a bowl $B$, {\it i.e.}, 
$\pddl{goal}$ is $\pddl{In}[A,B] = \texttt{True}$. 
To solve this, a class of TAMP solvers first finds a \emph{plan skeleton} $\vec{a}$, a sequence of actions with discrete arguments fixed, such as $\langle \pddl{pick}[A], \pddl{place}[A,B] \rangle$, and then grounds it into a \emph{factor graph} that connects its continuous parameters through its constraints (\autoref{fig:factor-graph}). 
A plan skeleton's factor graph $\Phi(\vec{a}) = (V, F)$ contains planning variables $V$, namely the state variables at every step of the skeleton, and action constraints as factors $F$. The skeleton is feasible if this constraint satisfaction problem (CSP) has an assignment $\beta$ of values to $V$ that satisfies every factor in $F$. %
Shared variables couple actions: in \autoref{fig:factor-graph}, the grasp $g_2$ appears in the \pddl{Kin} factors of both \pddl{pick} and \pddl{place}, so the grasp used to pick the apple must also allow a stable placement in the bowl. 
A number of TAMP algorithms perform constraint satisfaction by composing tools (also called generators~\cite{kaelbling2011hierarchical} or streams~\cite{garrett2020pddlstream}) into a \emph{computation graph}~\cite{garrett2020pddlstream, shen2025cutamp}, in which the outputs of earlier tool calls feed later ones.
There, a tool \emph{certifies} a factor when every value it outputs satisfies that factor.

\paragraph{Planning under uncertainty.} %
In the open world, the robot receives only indirect observations of $s_t$, making the problem a Partially Observable Markov Decision Process (POMDP)~\citep{kaelbling1998pomdp}. 
A common way to handle this added complexity is to combine planning with state estimation.
Namely, the robot maintains a \emph{belief} $b_t(s) = \Pr(s_t = s \mid h_t)$, a probability distribution over states, given its history $h_t$ of actions and observations, updated after every step.
From an initial belief $b_0$, the solution is a policy $\pi(a_t \mid b_t)$ that chooses an action $a_t$ from the belief $b_t$ and reaches a belief under which the goal holds with high probability. Belief-space TAMP~\citep{kaelbling2013belieftamp,garrett2020onlinereplanning} represents every plan variable by a belief, a distribution over its probable values. In the example, the apple's pose is no longer a known value $x_0$ but a belief $b_A$, the marginal of $b_t$ over $\pddl{at}[A]$, so the plan must first observe it: in \autoref{fig:factor-graph}, \pddl{detect}$[A]$ moves the robot to a viewpoint $q_1$ and turns $b_A$ into an observed pose $x_1$ through the \pddl{RGBD} factor. This uncertainty carries over to the grasp $g_2$, planned on $x_1$, and through $g_2$ to the placement $x_3$. Planning over beliefs requires (1) each constraint to hold with high probability under the belief and (2) the cost of each action to be weighed against its risk of failure~\citep{curtis2024tampura}, and both require integrating over the belief for every candidate plan. Belief-space TAMP therefore computes $\pi$ online: it plans from $b_t$, executes the first action, updates the belief, and replans~\citep{garrett2020onlinereplanning}.

\lstdefinestyle{typed}{
  basicstyle=\ttfamily\scriptsize, columns=fullflexible, keepspaces=true,
  backgroundcolor=\color{gray!10}, frame=single, xleftmargin=1em, xrightmargin=1em,
  emph={}, mathescape=true, escapeinside={},
  literate={pre:}{{\textbf{pre:}}}4 {eff:}{{\textbf{eff:}}}4 {con:}{{\textbf{con:}}}4 {types:}{{\textbf{types:}}}6
}

\section{TAMPAgent: Belief-Space Agentic TAMP}
\label{sec:method}

We want to build a method that allows an interface between the capability of planning under uncertainty using belief-space TAMP (\autoref{sec:preliminaries}) and the open-world agentic reasoning capabilities to solve open-world mobile-manipulation problems. In order to achieve this, we propose TAMPAgent, where we turn an LVLM \emph{agent} into a belief-space policy $\pi(a_t \mid b_t)$ for the planning problem of \autoref{sec:preliminaries}. The agent observes the scene, reasons in language, and calls perception, grasping, and motion tools, while the planner compiles its decisions into a constraint satisfaction problem~(CSP) and checks that all constraints are solved with certified values before anything executes. The two meet through hybrid beliefs, held in language until tools refine them to sufficient resolution for low-level action generation. In this section, we first define the agentic belief-space problem and then the TAMPAgent planning loop that solves it.

\begin{figure}[t]
    \centering
    \includegraphics[width=0.95\linewidth]{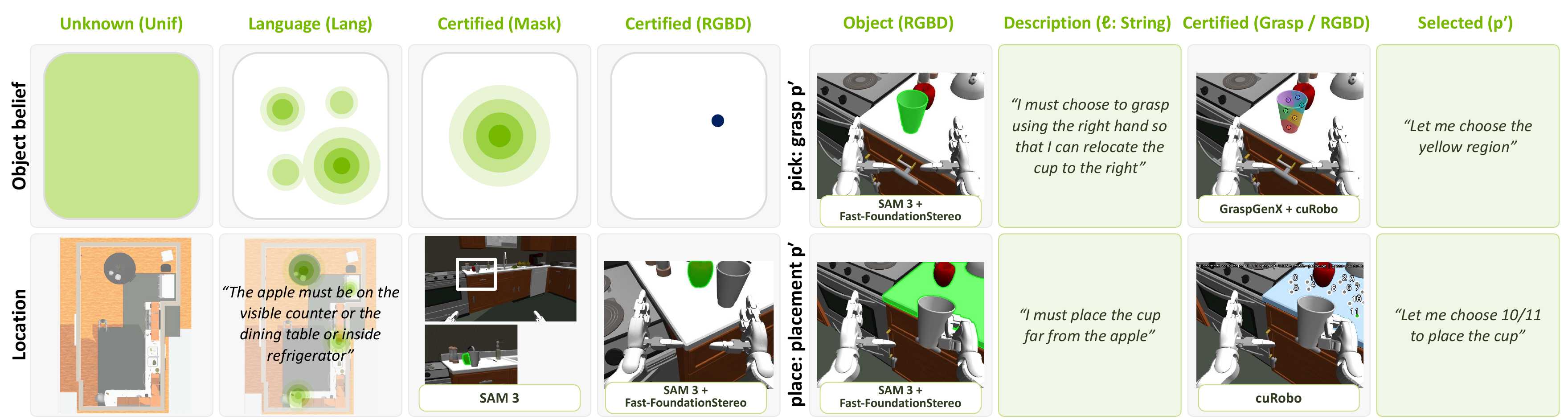}
    \caption{\textbf{Hybrid beliefs over perception and manipulation variables.}
\textit{Left:} The belief over the apple's location moves from $\mathsf{Unif}$, a uniform prior over the room, to a $\mathsf{Lang}$ hypothesis written by the agent, ``... on the visible counter / dining table / inside the refrigerator ...''. The agent then chooses a viewpoint and calls SAM~3 to represent apple's location as a $\mathsf{Mask}$, and then as a $\mathsf{RGBD}$ estimate when its close enough.
\textit{Right:} The grasp (top) and placement (bottom) $p'$ requires cup's $\mathsf{RGBD}$ estimate as input. The agent injects language context $\ell$, such as ``I must place the cup far from the apple'' and calls tools to certify valid grasps and placement locations. Finally, the agent chooses a valid option based on implicitly represented planning objectives. }
\label{fig:variable_ladder}
\vspace{-1em}
\end{figure}

\subsection{The Agentic Belief-Space Problem}
\label{subsec:formulation}

\textbf{Problem.} We leverage two useful qualities of agents: (1) the open-world understanding~\cite{kumar2024owltamp, shen2026tiptop} and (2) the capability to create plausible computation graphs using available geometric tools~\cite{chen2026gap}. At each step $t$, the agent instantiates the planning problem from its current history, context, and memory. It describes the objects it observes and writes a problem $P_t$ that assigns a belief to every state variable of these objects, together with their categories (for example, whether an object can be moved or can support others) and the goal parsed from the user instruction. As $P_t$ is rewritten at every step, newly discovered objects, such as a box occluding the apple, are added as soon as they are observed.

\textbf{Hybrid beliefs.} The agent reasons in language, but tools certify constraints only on continuous values. Each plan variable therefore holds a belief based on the available information (\autoref{fig:variable_ladder}): uniform prior ($\mathsf{Unif}$) when nothing much is known, a language hypothesis about where the probability mass lies ($\mathsf{Lang}$), or a value certified by a tool, such as a mask, an RGB-D estimate, a grasp, a configuration, or a trajectory. When a tool returns several certified values, the agent chooses one by objectives it represents implicitly, such as a grasp that suits a later placement. The language form acts as a belief statement~\citep{curtis2024tampura, kumar2024owltamp} with an open vocabulary written by the agent. Over it, the agent handles the chance constraints and costs of belief-space TAMP implicitly. A wrong judgment surfaces as a tool failure. Hence, every variable's belief is stored as language for the agent to read or as geometric certified values for the planner to check. 

\textbf{Actions over hybrid beliefs.} The domain has three actions: \pddl{detect}, \pddl{pick} and \pddl{place} defined below. Each action: (i) is characterized by the object $o$ it acts on and a description $\mathsf{String}$ $\ell$ that the agent writes from its context to explain the desired qualities of the outcome, (ii) accepts strict typed arguments by the form of their value and (iii) moves the robot along a trajectory $\tau$ to a configuration $q$ and changes the belief of $o$ from $p$ to $p'$. The available types include (a) Tool certified values: $\mathsf{Grasp}$, $\mathsf{Conf}$, $\mathsf{Traj}$, $\mathsf{Mask}$, and $\mathsf{RGBD}$, and (b) Agent written values: $\mathsf{Lang}$ and $\mathsf{String}$. The \pddl{detect} action moves the robot to a configuration from which the camera sees $o$, and refines its belief:
\begin{lstlisting}[style=typed]
types:   Grasp | Conf | Traj | Mask | RGBD | String | Lang | (Unif)

detect[o, $\ell$: String](p: Unif|Lang|Mask|RGBD,  p': Mask|RGBD,  q: Conf,  $\tau$: Traj)
  pre:  at[o] = p
  eff:  at[o] = p',  atRob = q
  con:  Motion(atRob, $\tau$, q),  Obs(o, q, p')
\end{lstlisting}
$\pddl{Obs}$ is shorthand for multiple observation modes representing the target object's pose using (a) $\pddl{Mask}(o, q, p')$ holds when $p'$ is the segmentation mask of $o$ seen from $q$, and (b) $\pddl{RGBD}(o, q, p')$ holds when $p'$ adds valid depth to a mask when the object is within manipulation reach. \autoref{fig:variable_ladder} (left) illustrates how \pddl{detect} moves a location belief through these forms: from uniform prior $\mathsf{Unif}$ to $\mathsf{RGBD}$ estimate that \pddl{pick} and \pddl{place} requires to manipulate an object:
\begin{lstlisting}[style=typed]
pick[o, $\ell$: String](p: RGBD,  p': Grasp,  q: Conf,  $\tau$: Traj)
  pre:  holding = None,  at[o] = p
  eff:  holding = o,  at[o] = p',  atRob = q
  con:  Grasp(o, p, p'),  Kin(o, p, p', q),  Motion(atRob, $\tau$, q)

place[o, o', $\ell$: String](p: Grasp,  p': RGBD,  p$_s$: RGBD,  q: Conf,  $\tau$: Traj)
  pre:  holding = o,  at[o] = p,  at[o'] = p$_s$
  eff:  holding = None,  at[o] = p',  On[o, o'] = True,  atRob = q   (In for a container)
  con:  Stable(o, o', p$_s$, p'),  Kin(o, p, p', q),  Motion(atRob, $\tau$, q)
\end{lstlisting}
\autoref{fig:variable_ladder} (right) shows the free parameter $p'$ of \pddl{pick} (top) and \pddl{place} (bottom). Each starts from the object's $\mathsf{RGBD}$ estimate and the agent written description $\ell$, the tools return certified grasps or placements, and the agent selects $p'$ among them as $\ell$ describes. After \pddl{pick}, $\pddl{at}[o]$ is the grasp, the object's pose in the hand, and \pddl{place} starts from it, so both \pddl{Kin} factors share the grasp, as in \autoref{fig:factor-graph}. The types enforce the order of actions: \pddl{pick} requires an $\mathsf{RGBD}$ belief, so it must follow a \pddl{detect} that produced one. The agent chooses valid grasps, placements, configurations, and trajectories among a tool's certified outputs. It proposes viewpoints directly: the base planner certifies that the robot can reach one (\pddl{Motion}), and once the robot is there, SAM~3 or the RGB-D estimate certifies that the object is visible from it (\pddl{Obs}).

\begin{figure}[t]
\centering
\includegraphics[width=\linewidth]{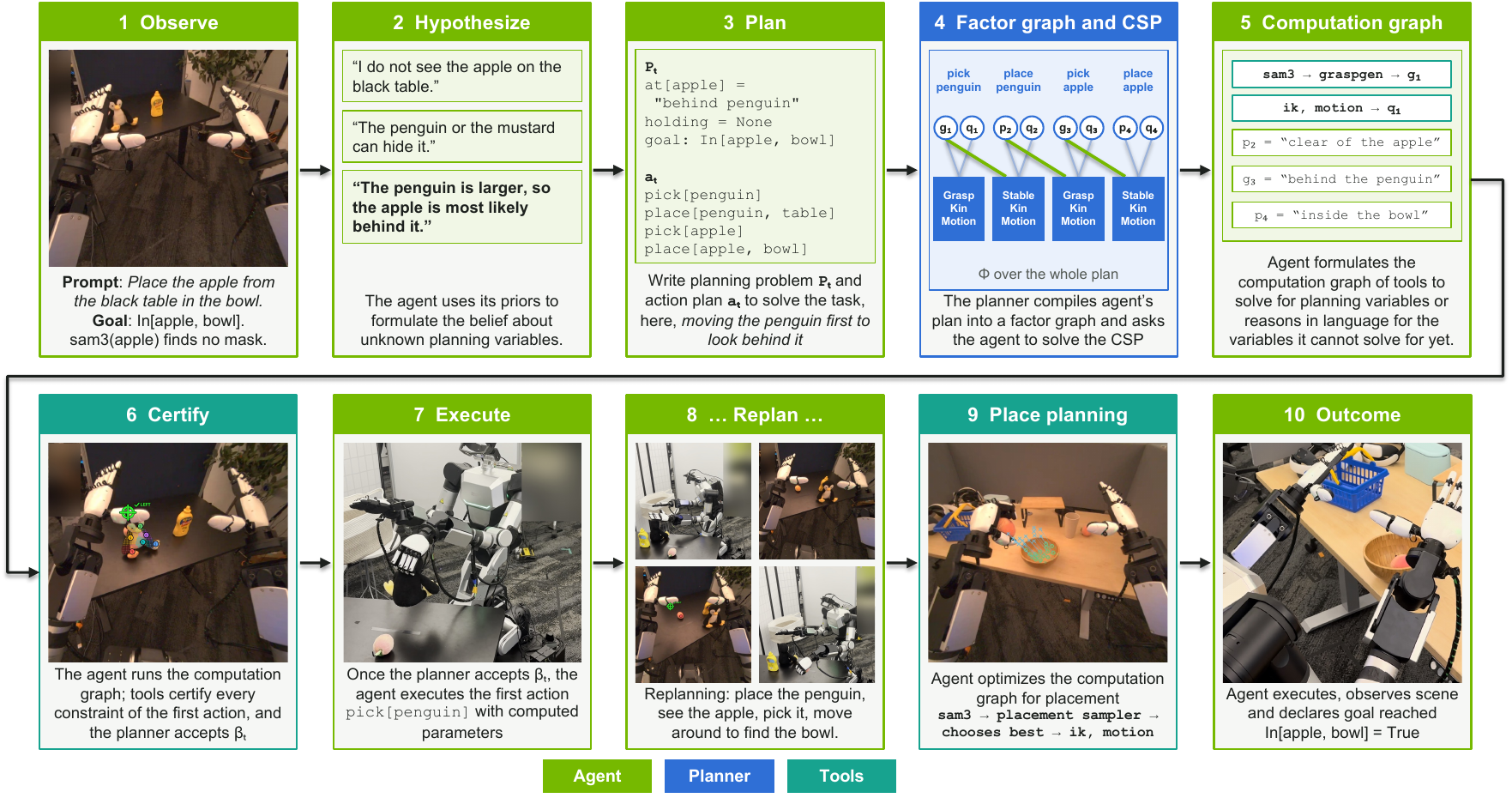}
\caption{\textbf{TAMPAgent planning loop.} We show how the agent orchestrates the planner and tools to solve the task ``put the apple on the black table (hidden behind a toy penguin) in a bowl''. (1-3) The agent observes, hypothesizes where the apple is, and writes the problem and a plan, (4) the planner compiles the plan into a CSP, and (5) the agent writes a computation graph that calls tools to find an assignment $\beta_t$ satisfying the CSP. (6) The tools certify the first action's constraints and the planner accepts $\beta_t$, allowing (7) the robot to execute the first action. (8) The loop repeats until (9-10) the placement is planned and the goal is achieved.
}
\vspace{-1em}
\label{fig:execution_loop}
\end{figure}

\textbf{Tools.} Each constraint is certified by a tool, which moves a belief from one form to the next:

\begin{center}
\small
\vspace{-2em}
\resizebox{\linewidth}{!}{%
\begin{tabular}{@{}lll@{}}
\toprule
Constraint & Certifying tool & Inputs $\rightarrow$ outputs \\
\midrule
\pddl{Mask} & SAM~3~\citep{carion2025sam3segmentconcepts} & language belief, viewpoint $\rightarrow$ mask \\
\pddl{RGBD} & SAM~3 and Fast-FoundationStereo~\citep{wen2026fast} & mask, nearby viewpoint $\rightarrow$ RGB-D estimate \\
\pddl{Grasp} & GraspGenX~\citep{han2026graspgenx} & RGB-D estimate $\rightarrow$ set of valid grasps \\
\pddl{Kin} & cuRobo inverse kinematics~\citep{sundaralingam2026curobov2} & RGB-D estimate, grasp $\rightarrow$ configuration \\
\pddl{Motion} & cuRobo motion planner; base motion planner & configurations $\rightarrow$ arm or base trajectory \\
\pddl{Stable} & placement sampler; cuRobo for IK and motion planning & support, placement hypothesis $\rightarrow$ placement poses \\
\bottomrule
\end{tabular}}
\end{center}

\noindent Compiling a plan therefore points towards the tools required to satisfy the constraints and also inform the agent if certain variables are shared. A tool failure informs the agent about an unsatisfied constraint softly-guiding the agent to revise the strategy. Adding a capability to the framework  amounts to adding a tool with the constraint it certifies.

\subsection{TAMPAgent Planning Loop}
\label{subsec:policy}

\textbf{Stages.} The TAMPAgent planning loop, as illustrated in \autoref{fig:execution_loop}, has three stages. (i) The agent observes, hypothesizes, and writes $P_t$. The planner grounds the domain's actions over the objects and categories in $P_t$ using standard task planning techniques~\citep{garrett2021integratedtamp}. The agent then composes candidate plans from these grounded actions, and chooses a suitable $\vec{a}_t$ based on its priors (panels 1 to 3). (ii) The planner compiles $\vec{a}_t$ into the CSP $\Phi(\vec{a}_t)$ (panel 4). (iii) The agent writes the computation graph $G_t$ of tools to bind the first action's arguments, keeping later variables in language (panel 5). Once the tools certify every constraint of the first action, the planner accepts $\beta_t$ (panel 6), the robot executes that action (panel 7), and the agent replans from the new observation (panel 8) until the placement is planned and the goal is achieved (panels 9 and 10). During this process, the agent also reasons about coupled variables, for example, it selects the grasp $g_2$ to suit a placement that is still a language hypothesis. We allow the agent to write and execute $G_t$ as raw Python scripts in a sandbox (\autoref{app:sandbox}) such that the agent's priors enter through its prompts, viewpoints, and choices among candidates. Unlike agents that write one graph for a whole task and validate it by rehearsal~\citep{chen2026gap}, TAMPAgent writes $G_t$ at each step against the constraints the planner declares.
 
\textbf{Illustrative example.} We show the first two actions of the skeleton in \autoref{fig:factor-graph}, \pddl{detect}$[A]$ and \pddl{pick}$[A]$, with goal $\pddl{holding} = A$. Each block is one pass through the loop, and its rows follow the stages: the problem $P_t$, plan $\vec{a}_t$, constraints of the first action ($\Phi[{:}1]$) and the later ones ($\Phi[1{:}]$), the computation graph $G_t$, accepted assignment $\beta_t$, and the executed action $a_t$ along with its effect. 
 
\begin{stepbox}{typed}{Step 0: observe the apple}
$P_0$          at[A]: Lang = "possibly on the brown table", holding = None, atRob = q$_0$, goal: holding = A
(*@\steprule@*)$\vec{a}_0$          detect[A, "facing the brown table"] -> pick[A, "top grasp, clear of the table edge"]
(*@\steprule@*)$\Phi$[:1]       Motion(q$_0$, $\tau_1$, q$_1$),  Obs(A, q$_1$, x$_1$)
$\Phi$[1:]       Grasp(A, x$_1$, g$_2$),  Kin(A, x$_1$, g$_2$, q$_2$),  Motion(q$_1$, $\tau_2$, q$_2$)
(*@\steprule@*)$G_0$          q$_1$: Conf = agent.choose_viewpoint("facing the brown table")
             $\tau_1$: Traj = base_planner.plan(q$_0$, q$_1$)          # certifies Motion
(*@\steprule@*)$\beta_0$          {q$_1$, $\tau_1$} accepted
(*@\steprule@*)$a_0$          detect[A, "facing the brown table"](b$_A$, x$_1$, q$_1$, $\tau_1$) -> at[A]: RGBD = x$_1$,  atRob = q$_1$
\end{stepbox}
\pddl{pick}'s constraints $\Phi[1{:}]$ are deferred because they depend on $x_1$, which is yet to be determined. After execution, \pddl{detect} turns the apple's belief 
into $\mathsf{RGBD}$, enabling the agent to plan \pddl{pick} next.
 
\begin{stepbox}{typed}{Step 1: pick the apple}
$P_1$          at[A]: RGBD = x$_1$,  holding = None,  atRob = q$_1$,  goal: holding = A
(*@\steprule@*)$\vec{a}_1$          pick[A, "top grasp, clear of the table edge"]
(*@\steprule@*)$\Phi$[:1]       Grasp(A, x$_1$, g$_2$),  Kin(A, x$_1$, g$_2$, q$_2$),  Motion(q$_1$, $\tau_2$, q$_2$)
(*@\steprule@*)$G_1$          grasps = graspgenx.generate(x$_1$)                 # certifies Grasp
             g$_2$: Grasp = agent.choose(grasps, "top grasp, clear of the table edge")
             q$_2$: Conf = curobo.ik(x$_1$, g$_2$)                       # certifies Kin
             $\tau_2$: Traj = curobo.plan_motion(q$_1$, q$_2$)           # certifies Motion
(*@\steprule@*)$\beta_1$          {g$_2$, q$_2$, $\tau_2$} accepted
(*@\steprule@*)$a_1$          pick[A, "top grasp, ..."](x$_1$, g$_2$, q$_2$, $\tau_2$) -> holding = A,  at[A] = g$_2$,  atRob = q$_2$
\end{stepbox}
The grasp passes through two forms: GraspGenX~\cite{han2026graspgenx} with motion planning using cuRobo~\cite{sundaralingam2026curobov2} returns candidates grasps, and the agent chooses one based on its language context $\ell$. Thus, only certified values reach the robot. The grasp $g_2$ selected here is shared with the \pddl{Kin} factor of \pddl{place}~(\autoref{fig:factor-graph}), so the agent commits to it while considering the placement $x_3$ as $\mathsf{Lang}$.

\textbf{Acceptance and recovery.} The planner accepts $\beta_t$ once every constraint of the first action is certified and every argument has the form its type requires. As all arguments that the agent uses to execute low-level control are tool certified under current robot state and perception, the agent can  pursue a poor plan but not an infeasible action (assuming all tools are correct). As the robot state changes after execution, the agent needs to rewrite the problem and replan, reusing the rest of $\vec{a}_t$ where possible~\citep{garrett2020onlinereplanning}. When no computation graph certifies the first action, the agent escalates through (1) other candidates or viewpoints, (2) another plan, or (3) new hypotheses. This resolves both (a) \emph{epistemic gaps}, such as a box occluding the apple, and (b) \emph{kinematic gaps}, such as a cup blocking every grasp, by extending the plan to move the blocker before any motor execution. \autoref{app:trace} traces the full run of \autoref{fig:execution_loop}, including both gaps.
\section{Evaluation}
\label{sec:benchmark-results}

\begin{figure}[t]
\centering
\includegraphics[width=\linewidth]{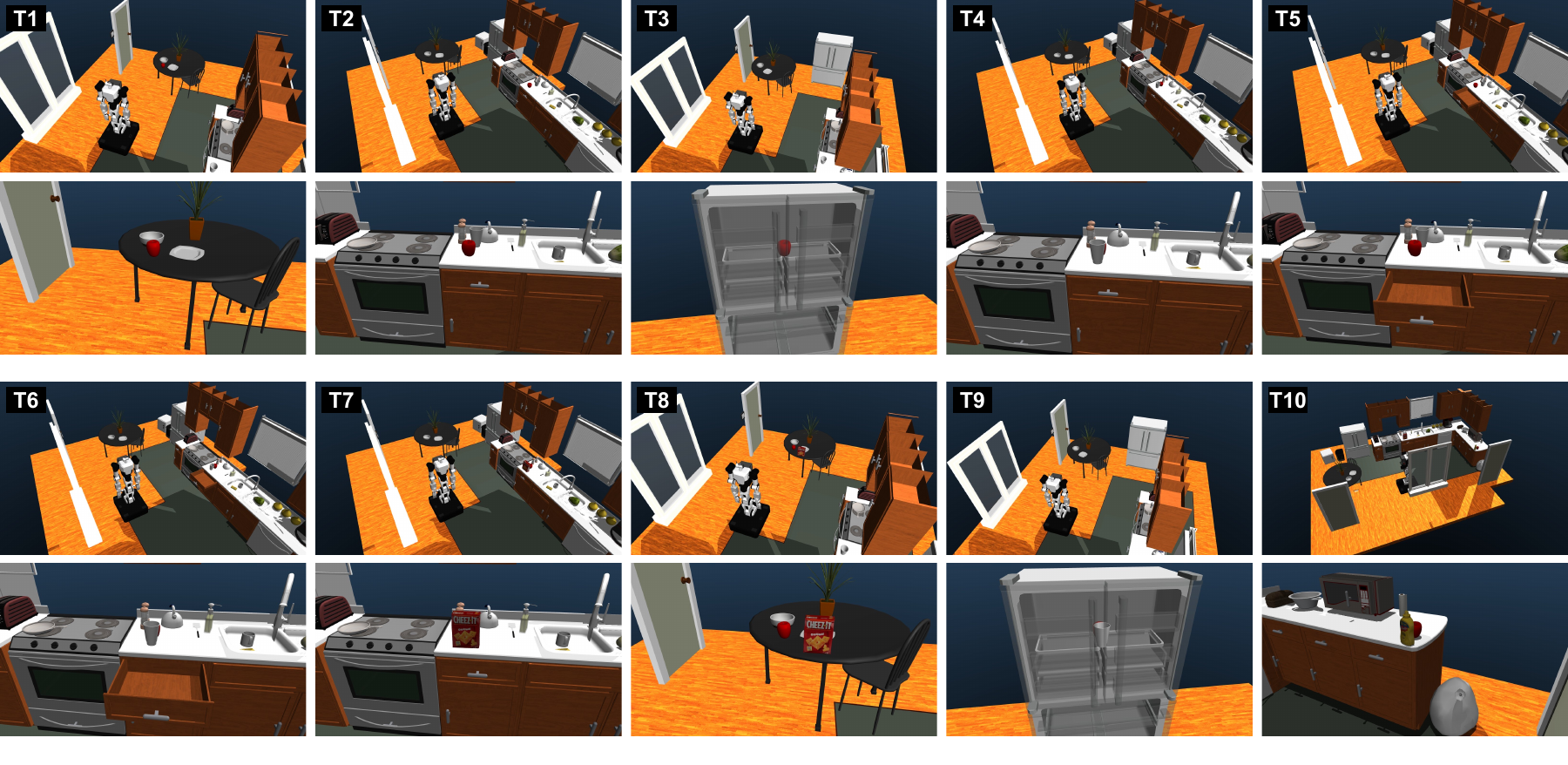}
\caption{\textbf{Evaluation tasks.} We ask the robot to place the apple on the plate on the dining table (prompts in \autoref{tab:task-prompts}), starting with the robot facing away from the apple. Each task shows the whole scene with the robot at its start pose (top, walls and ceiling hidden) and a close-up of the area around the apple (bottom). The tasks fall into four categories. In the mobile manipulation (MM) tasks T1 and T2, the apple is in plain view, next to the plate or on the countertop. The MM + non-prehensile tasks add an object the robot must move first. In T4 to T6, a cup, an open drawer, or both block the direct pick, in T10, a mustard bottle stands in front of the apple, and in T8, a cracker box stands in front of the plate. The MM + occlusion tasks hide the apple, inside the fridge in T3 and behind a cracker box in T7, so the agent must hypothesize where to search. T9 is MM + non-prehensile + occlusion: the apple is inside the fridge behind a cup, which the robot must move before it can pick the apple. In the close-ups of T3 and T9, we draw the refrigerator doors translucent so that the shelf is visible. In the simulation the doors are closed.}
\label{fig:sim-tasks}
\end{figure}

\begin{figure}[t]
\centering
\includegraphics[width=\linewidth]{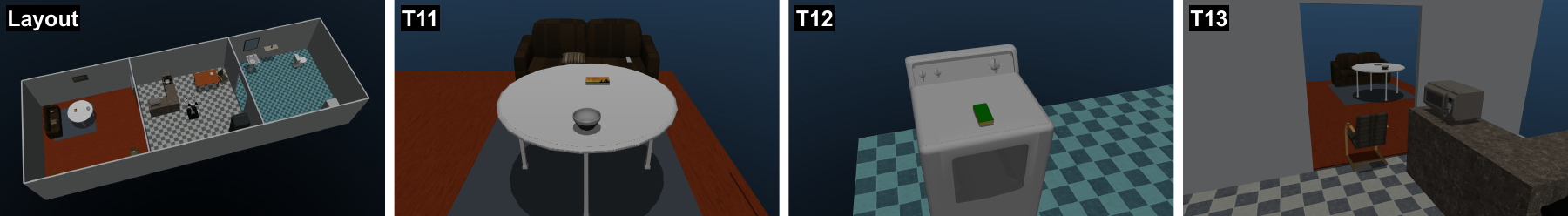}
\caption{\textbf{Three-room tasks.} T11 to T13 take place in a three-room home with a living room, a kitchen, and a washroom. In T11, the robot must move a bowl from the living-room table to the kitchen table, and in T12, it must throw a used sponge from the laundry machine in the washroom into the kitchen trash can. T13 has the same goal as T11, but a chair blocks the living-room doorway. We show the layout of the home at the start of T11 (ceiling hidden), followed by close-ups of the bowl on the living-room table (T11), the sponge on the laundry machine (T12), and the chair in the living-room doorway, seen from the kitchen (T13).}
\label{fig:multiroom-tasks}
\end{figure}

We evaluate the impact of endowing an agent with a belief-space planning model by comparing TAMPAgent with two baselines. In Code as Policies (CaP)~\citep{liang2023codeaspolicies}, the agent writes code that calls the tools and decides when to move. In Graph as Policy (GaP)~\citep{chen2026gap}, helper agents author and review graphs of tool calls, and the robot moves only through a validated graph (\autoref{app:baselines}). All methods use Codex~\citep{bolin2026codexagentloop} running GPT-6-Astra~\citep{openai2026gpt6astra} at low reasoning effort on the RBY1 (GPT-5.6 Sol~\citep{openai2026gpt56sol} at medium reasoning effort on the G1, \autoref{app:g1}), with the same tools and documentation (\autoref{sec:method}). We test three hypotheses:
\textbf{(H1)} TAMPAgent promotes hypothesis-driven reasoning to leverage the agent's priors for reducing uncertainty.
\textbf{(H2)} TAMPAgent promotes reasoning about inter-step dependencies, geometric reachability, and affordance across extended horizons.
\textbf{(H3)} With the same tools and documentation, the belief-space planning representation allows the agent to efficiently reason using a compact, structured view of the task, so it uses fewer tokens than when it uses the tools directly.

\textbf{Tasks and metrics.} Every kitchen task gives a short goal prompt (\autoref{tab:task-prompts}), starts with the robot facing away from the apple, and provides no scene description as illustrated in~\autoref{fig:sim-tasks}. We simulate an RBY1 mobile manipulator and a G1 humanoid in MuJoCo~\citep{todorov2012mujoco}, in a kitchen built from MolmoSpaces~\citep{kim2026molmospaces} and RoboCasa~\citep{robocasa365} assets, and run 20 episodes per task and method on the RBY1 (\autoref{app:tasks}), except T9, on which we measure learning from attempts, and 5 episodes on T1 to T6 on the G1~(details in~\autoref{app:g1}). The kitchen tasks fall into four categories. \emph{Mobile manipulation} (MM: T1, T2) requires finding the apple and the plate, picking, carrying, and placing. \emph{MM + non-prehensile} (T4 to T6, T8, T10) adds an object or a drawer that blocks the pick or the place, which the robot must move or close first. \emph{MM + occlusion} (T3, T7) hides the apple inside the fridge or behind an object, so the agent must hypothesize where it is and uncover it. \emph{MM + non-prehensile + occlusion} (T9) hides the apple inside the fridge behind a cup that blocks the pick. T11 to T13 take place in a three-room home (\autoref{fig:multiroom-tasks}), where the robot must move a bowl from the living room to the kitchen table (T11, T13) or trash a used sponge from the washroom in the kitchen trash can (T12). For these three tasks, we run 5 episodes per task and method on the RBY1. An episode succeeds when the apple is on the plate, and in T11 to T13, when the bowl is on the kitchen table or the sponge is in the trash can. It fails when an object falls or the scene is destroyed, or when its budget of simulated time runs out, 15 minutes plus 5 minutes for each blocker, and 25 minutes for T3, T7, T9, and T10. We report (i) absolute success, (ii) partial success, the fraction of applicable subtasks completed (find the apple, clear each blocker, pick, find the plate, place), (iii) the cause of each failure, and (iv) uncached input and output tokens.

\begin{figure}[t]
\centering
\includegraphics[width=\linewidth]{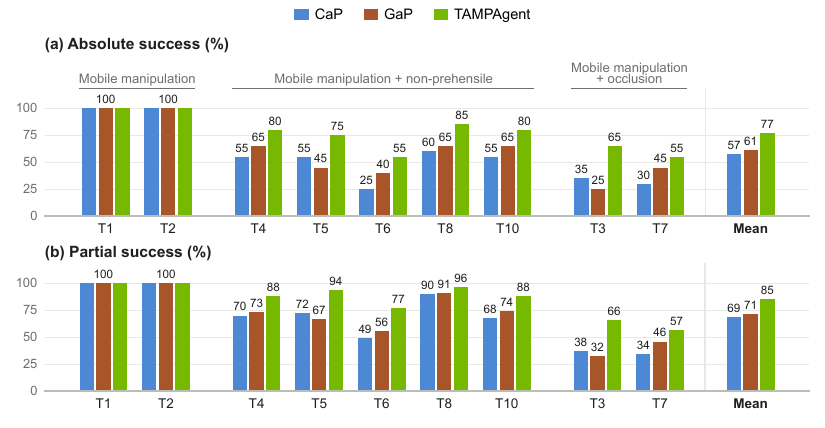}
\caption{\textbf{Success on the RBY1}, over 20 episodes per task and method, with the tasks grouped by category. (a) Absolute success. (b) Partial success, where each episode earns the fraction of its subtasks it completes (\autoref{fig:rby1-subtasks}, \autoref{app:rby1-subtasks}). The last group averages each over the tasks.}
\label{fig:rby1-success}
\vspace{-0.8em}
\end{figure}

\begin{figure}[t]
\centering
\includegraphics[width=\linewidth]{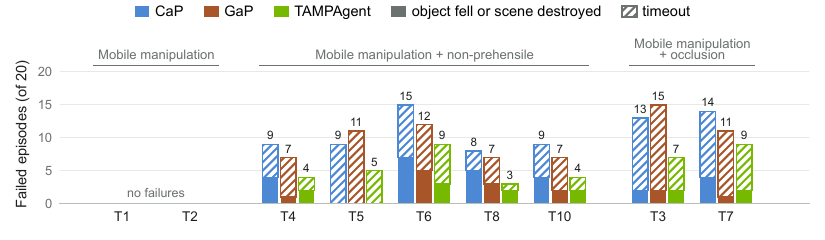}
\caption{\textbf{Failure attribution for RBY1 simulation experiments.} Failed episodes out of 20 per task, in which an object fell or the scene was destroyed (solid), or which timed out (hatched).}
\label{fig:rby1-failures}
\vspace{-0.8em}
\end{figure}

\begin{wrapfigure}{r}{0.34\linewidth}
\centering
\vspace{-1em}
\includegraphics[width=\linewidth]{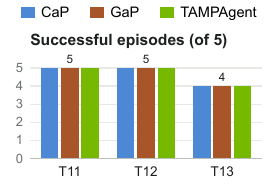}
\caption{\textbf{Success on the three-room tasks} (\autoref{fig:multiroom-tasks}) on the RBY1, over 5 episodes per task and method.}
\label{fig:multiroom-success}
\vspace{-1em}
\end{wrapfigure}

\textbf{H1: TAMPAgent promotes hypothesis-driven reasoning to leverage the agent's priors for reducing uncertainty.} In TAMPAgent, writing $P_t$ asks the agent to state a language belief over the location of each object it needs, so it lays out several hypotheses at once and ranks them by its priors. On the contrary, CaP searches one location at a time and proposes the next only after the current one fails. On T1 and T2, the apple is in plain view, so there are no competing hypotheses to rank, and all three methods succeed in every episode (\autoref{fig:rby1-success}). In the occlusion tasks, the apple is inside the fridge (T3) or behind a cracker box (T7). TAMPAgent includes the fridge or the box among its first hypotheses and plans to look there, so it finds the apple in 14 and 13 of 20 episodes, against 9 and 10 for both CaP and GaP (\autoref{fig:rby1-subtasks}, \autoref{app:rby1-subtasks}). CaP's sequential search consumes its time budget, and most failures of CaP and GaP on these two tasks are timeouts (\autoref{fig:rby1-failures}). In the three-room home, CaP, GaP, and TAMPAgent each succeed in 5 of 5 episodes on T11 and T12 and in 4 of 5 on T13 (\autoref{fig:multiroom-success}).

\textbf{H2: TAMPAgent promotes geometric reasoning about inter-step dependencies and affordance across extended horizons.}
As the planner provides the agent with a set of grounded actions based on the problem it writes, actions that move a blocker are among the agent's candidate plans from the start. When the direct pick-place plan fails certification, the failure names the blocking factor, and the agent extends the plan with suitable actions. The non-prehensile tasks T4 to T6, T8, and T10 block the direct plan, and here TAMPAgent succeeds in 75\% of episodes, against 50\% for CaP and 56\% for GaP (\autoref{fig:rby1-success}). In T4, the cup blocks every grasp of the apple, and the plan must place the cup where the later \pddl{pick} of the apple remains feasible: where the cup goes is a variable of its \pddl{place}, but whether that placement is acceptable is decided by the \pddl{Kin} and \pddl{Motion} factors of that \pddl{pick}. In T5, the open drawer makes the \pddl{Motion} factor of \pddl{detect} unsatisfiable, so the planner finds the collision before the robot moves, and closing the drawer becomes a subgoal. T6 adds the order in which to clear both blockers. In T10, the mustard bottle must be moved before the apple can be picked, and in T8, the box in front of the plate restricts where the apple can be placed. CaP has to come up with clearing the blocker itself and checks no constraint before acting. On these five tasks, CaP, GaP, and TAMPAgent drop an object or destroy the scene in 20, 11, and 9 episodes, and time out in 30, 33, and 16 (\autoref{fig:rby1-failures}).

\textbf{H3: Belief-space planning representation allows the agent to reason and act efficiently with a compact, structured view of the task.} All three methods receive the same tools and documentation, yet per minute of simulated time TAMPAgent uses 19.3k uncached input and 1.75k output tokens, against 20.3k and 1.91k for CaP and 27.4k and 2.73k for GaP. We normalize by simulated time because episode lengths differ. The representation saves tokens in two ways. On the input side, $P_t$ is a compact state the agent can read. It records the known objects and the certified beliefs, so the agent needs to consult the documentation less often than CaP. On the output side, $\Phi(\vec{a}_t)$ names the constraints of the next action and the tool that certifies each, so the agent writes $G_t$ as a short composition of those calls. CaP must decide in its program which tools to call and when the robot moves, and GaP, though more structured, must come up with the whole graph by itself.

\begin{figure}[h]
\centering
\includegraphics[width=0.9\linewidth]{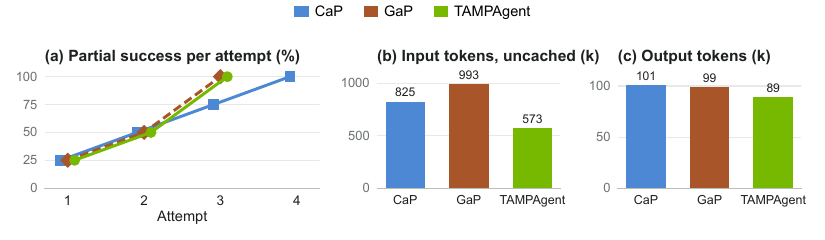}
\caption{\textbf{Learning from attempts on T9.} (a) Partial success of each attempt, the fraction of the four stages (approach the refrigerator and find the apple, relocate the cup, pick the apple, place the apple) it completes, until the first success. (b) Uncached input and (c) output tokens, summed over all attempts. The tokens of GaP include its helper agents.}
\label{fig:t9-learning}
\vspace{-0.8em}
\end{figure}

\textbf{Learning from attempts.} In T9, the apple is inside the refrigerator blocked by the cup. Solving the task with generic goal prompt ``place apple on plate on dining table'' requires: (i) finding the apple on supports and containers, (ii) clearing the blocker within the narrow passage of the refrigerator, and (iii) picking and placing the apple. For this experiment, we remove all strategical context from the agent-helper-docs, let the agent attempt the task repeatedly in the simulator, save memory context and learnings, and use it to perform better in subsequent trials. We evaluate CaP, GaP and TAMPAgent based on number of attempts required to solve the task and the total tokens used in the process. We divide the task into four stages: approaching the refrigerator, which the robot completes when it opens the door and finds the apple, relocating the cup out of the way of the apple, picking the apple, and placing it on the plate. The partial success of an attempt is the fraction of these stages it completes. TAMPAgent and GaP succeed in their third attempt and CaP in its fourth (\autoref{fig:t9-learning}a), and over these attempts TAMPAgent uses 42\% and 31\% fewer uncached input tokens than GaP and CaP, and 9\% and 12\% fewer output tokens (\autoref{fig:t9-learning}b, c). This supports H3 and validates that TAMPAgent's representation results in more efficient recursive self-improvement.

\begin{figure}[h]
\centering
\includegraphics[width=\linewidth]{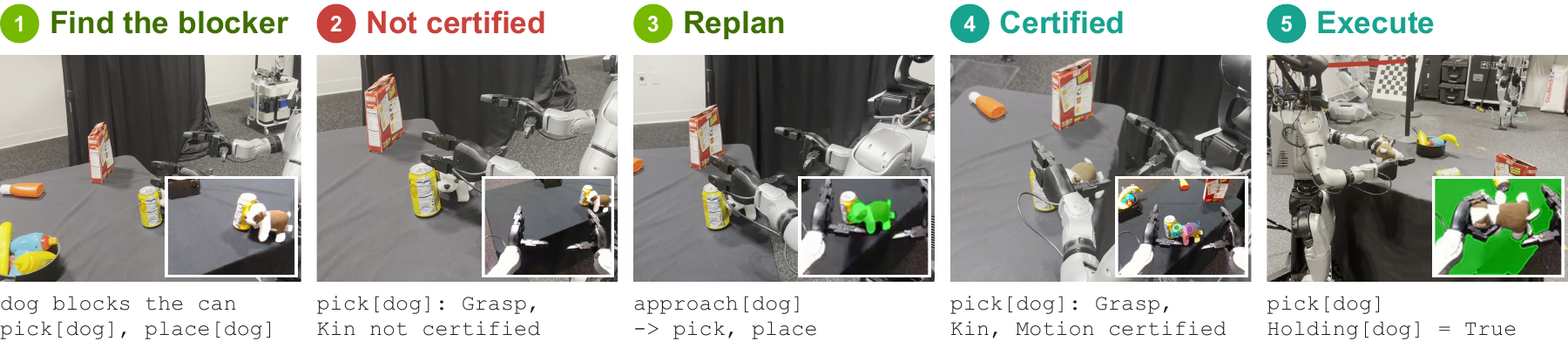}
\caption{\textbf{Clearing a blocker on a real G1.} The goal is to hold a yellow can. (1) The agent finds a toy dog that blocks the grasp of the can and plans to move it. (2) From the first stance, the planner does not certify the grasp and kinematics of \pddl{pick[dog]}, so (3) the agent adds an approach to the plan. (4) From the new stance, the tools certify the grasp, kinematics, and motion, and (5) the robot picks and lifts the dog. Insets show the robot's camera, the SAM~3 masks, and the grasp candidates. The run ends before the can is picked, when placing the dog is not certified.}
\label{fig:g1-real}
\end{figure}

\textbf{Real-world experiments.} We demoed TAMPAgent on a real RBY1 
(\autoref{app:hardware}) on three tasks: (a) \emph{simple pick and place}, (b) \emph{out-of-view receptacle}, in which the robot carries the object while it searches for the receptacle, and (c) \emph{hidden object}, in which the agent must hypothesize which object hides the target and move it, as in \autoref{fig:execution_loop}. 
Uncached I/O token usage per episode grows from 376k/39k tokens for (a) to 438k/58k for (b) and 973k/127k for (c). 
On a real G1, the agent clears a toy dog that blocks the grasp of a can (\autoref{fig:g1-real}). Videos are on our website, \url{https://umishra.me/TAMPAgent/}.

\section{Related Work}
\label{sec:related-work}

\textbf{Task planners and geometric tools.} Classical TAMP couples symbolic and geometric constraint satisfaction~\citep{garrett2020pddlstream,garrett2021integratedtamp}. Belief-space variants of TAMP plan over state distributions and replan upon observation~\citep{kaelbling1998pomdp,kaelbling2013belieftamp,garrett2020onlinereplanning}, extending to POMDPs~\citep{curtis2024tampura,phiquepal2022pto,tang2026popddl}. Others persist out-of-view states~\citep{liu2025dynamem,li2026soma,kim2026serf,chung2026liberomem,anwar2025remembr,sridhar2026memer,mark2026bpp,torne2026mem,lei2026robomemarena} over mapping substrates~\citep{wu2023objectslam,campos2021orbslam3,hornung2013octomap}. Mobile platforms add continuous loco-manipulation coupling~\citep{shamsah2023leggedbelief,ewen2021stormmap,ciebielski2025humanoidtamp,honerkamp2024momallm,shah2025bumble,liu2024okrobot,fu2024mobilealoha}. However, these systems require predefined workspaces and vocabularies. Uncertainty applies only to known object states, and predicates and operators stay fixed. Classical TAMP can therefore verify geometry but cannot generate the objects, predicates, and operators an unmapped environment requires.

\textbf{Agents in the planning loop.} Recent work uses LVLMs for open-world reasoning and differs in where the control loop resides. In the CaP paradigm, agents use tools directly: models bypass any structured solver and emit programs that call perception and control~\citep{liang2023codeaspolicies,singh2022progprompt,huang2023voxposer,ichter2023saycan,huang2023innermonologue,ren2023knowno,huang2025rekep,wu2023tidybot,ahn2024autort,shah2025bumble}. BUMBLE~\citep{shah2025bumble}, for instance, scales VLM skill selection to building-wide mobile manipulation, but it has no planner and checks no constraint before the robot acts. Such free-form agents often hallucinate skills because nothing tests their programs geometrically~\citep{chen2026gap,fu2026capx,chen2026volo,galanti2026addressing,lu2026aspire,xiao2026enpire,liu2026goal2skill,lei2026vlastools,qu2025pragmabot,shao2025symskill}. In the VLM-TAMP paradigm, a structured task planner replaces free-form programs and calls the agent and the tools. This restricts the agent to the planner's vocabulary~\citep{kumar2024owltamp,shen2026tiptop,chen2023autotamp,ding2023llmtamp} and uses it only for search heuristics~\citep{yang2024vlmtamp,wang2024llm3,joublin2024copal} or uncertainty estimation~\citep{zhao2025seeing,kim2026cocotamp,xiao2026hypothesis,jauhri2023activeperceptive,naveen2026languageassensor}. OWL-TAMP~\citep{kumar2024owltamp} has the model generate constraints and partial plans that a TAMP solver completes, assuming the relevant objects are already observed. Seeing is Believing~\citep{zhao2025seeing} uses foundation models as uncertainty estimators for a belief-space planner, which remains the policy. TiPToP~\citep{shen2026tiptop} queries the model once and then executes open loop. Existing methods thus either let the agent act without a verifier or confine it to a fixed solver pipeline. TAMPAgent lets the agent orchestrate both the planner and the tools, and the planner decides which constraints the tools must certify before each action executes.
\section{Conclusion}
\label{sec:conclusion}

We presented TAMPAgent, a framework that integrates LVLM agents and belief-space task and motion planning for mobile manipulation.
The agent acts as the belief-space policy over hybrid beliefs, and a planner compiles its plans into constraints that tools must certify before any action executes. In simulation on two robots, one of which is a humanoid, TAMPAgent 
outperforms CaP, and GaP on the RBY1, in kitchen tasks that require hypotheses and dependent steps, while using fewer tokens.

\section{Limitations and AI Usage Disclosure}
\label{app:limitations}

\textbf{Limitations.} TAMPAgent lets the agent propose any plan expressible in its domain and executes an action only once tools certify its constraints, so its guarantees are only as strong as those tools and the robot's perception. Because the agent weighs constraints and costs without calibrated probabilities, it sometimes keeps searching where acting would succeed, which causes most of its timeouts. The domain is fixed, so the agent cannot add actions on its own, and certification protects geometric feasibility but not the agent's interpretation of the task. 

\textbf{AI usage disclosure.}
Apart from the agentic nature of the proposed method, we used AI tools for writing assistance (grammar, style, editing, and paraphrasing), code generation and debugging, and generating figures of the experimental results.

\bibliography{iclr2026_conference}

\clearpage
\appendix
\setcounter{figure}{0}\renewcommand{\thefigure}{A\arabic{figure}}
\setcounter{table}{0}\renewcommand{\thetable}{A\arabic{table}}
\etocdepthtag.toc{mtappendix}
\etocsettagdepth{mtmain}{none}
\etocsettagdepth{mtappendix}{subsection}
\etocsettocstyle{\section*{Appendix Contents}}{}
\tableofcontents
\clearpage

\ifdefined\ArxivMainText\else
\section{Limitations and AI Usage Disclosure}
\label{app:limitations}

\textbf{Limitations.} TAMPAgent lets the agent propose any plan expressible in its domain and executes an action only once tools certify its constraints, so its guarantees are only as strong as those tools and the robot's perception. Because the agent weighs constraints and costs without calibrated probabilities, it sometimes keeps searching where acting would succeed, which causes most of its timeouts. The domain is fixed, so the agent cannot add actions on its own, and certification protects geometric feasibility but not the agent's interpretation of the task. 

\textbf{AI usage disclosure.}
Apart from the agentic nature of the proposed method, we used AI tools for writing assistance (grammar, style, editing, and paraphrasing), code generation and debugging, and generating figures of the experimental results.

\fi

\section{Detailed Method Explanation}
\label{app:running-example}

This section follows one real run from instruction to goal, in the notation of \autoref{sec:method}. Each step lists the problem $P_t$, the plan $\vec{a}_t$, the constraints $\Phi[{:}1]$ of its first action, the computation graph $G_t$, the accepted assignment $\beta_t$, and the executed action $a_t$ with its effects. Explanations follow the steps that introduce a part of the method. The run contains both gaps of \autoref{subsec:policy}: an epistemic gap, where perception cannot certify where the target is, and a kinematic gap, where no tool can certify a motion that achieves the next action. Tool outputs are those recorded in the run. A failure is labeled \texttt{OCCLUDED} when the target cannot be certified from the viewpoint, and \texttt{EMPTY} when a tool returns no value that satisfies its factor.

\subsection{Setup}
\label{app:domain}

The instruction is \emph{``place the apple from the black table in the wooden bowl''}. A black table holds the apple, a mustard bottle, and a toy penguin that hides most of the apple from every view the robot can reach. A wooden bowl stands on a desk elsewhere in the room. The robot starts facing away from both.

The domain is the one of \autoref{subsec:formulation}, with the actions \pddl{detect}, \pddl{pick}, and \pddl{place}. Two of its properties matter throughout the run. Because \pddl{pick} sets $\pddl{at}[o]$ to the grasp and \pddl{place} starts from it, the grasp chosen for a pick reappears in the \pddl{Kin} factor of the following place, the coupling of \autoref{fig:factor-graph}. A robot that holds an object can still execute \pddl{detect}, so it can search for the bowl while carrying the apple. We write $x$ for poses, $g$ for grasps, and $x'$ for placements, subscripted by object ($A$ apple, $T$ toy penguin, $B$ bowl, $S$ black table), and $q_{t+1}$ and $\tau_{t+1}$ for the configuration and trajectory that the action of step $t$ moves to.

\subsection{The Run, Step by Step}
\label{app:trace}

The agent writes the problem with both locations as language beliefs and plans the direct sequence.
\begin{stepbox}{typed}{Step 0: the target is not where the agent expects}
$P_0$          movable[apple] = True,  container[bowl] = True,  support[table] = True
             at[apple]: Lang = "on the black table",  at[bowl]: Lang = "on the wooden desk"
             holding = None,  atRob = q$_0$,  goal: In[apple, bowl] = True
(*@\steprule@*)$\vec{a}_0$          detect[apple, "facing the black table"]
             -> pick[apple, "a grasp that lets me drop it into the bowl"]
             -> detect[bowl, "toward the wooden desk"] -> place[apple, bowl, "inside the bowl"]
(*@\steprule@*)$\Phi$[:1]       Motion(q$_0$, $\tau_1$, q$_1$),  Obs(apple, q$_1$, x$_A$)
(*@\steprule@*)$G_0$          q$_1$: Conf = agent.choose_viewpoint("facing the black table")
             $\tau_1$: Traj = base_planner.plan(q$_0$, q$_1$)                  # certifies Motion
(*@\steprule@*)$\beta_0$          {q$_1$, $\tau_1$} accepted
(*@\steprule@*)$a_0$          detect[apple, "facing the black table"](b$_A$, x$_A$, q$_1$, $\tau_1$)
             sam3("apple") from q$_1$ -> 303 px mask, 1 depth px
             FAILURE  Obs of detect[apple]: OCCLUDED;  at[apple] unchanged,  atRob = q$_1$
\end{stepbox}

\textbf{The epistemic gap.} The robot cannot certify the apple's location from $q_1$. The penguin hides most of the apple, so the mask is small and has almost no depth, and another viewpoint of the same table would return the same answer. The failure is attached to the observation constraint of \pddl{detect}, so the agent revises its belief about the apple. TAMPAgent closes the gap with the agent's priors: the agent explains the failure with a hypothesis about where the apple can be, and extends the plan with actions that would make that hypothesis observable.

\begin{stepbox}{typed}{Step 1: hypothesize, plan, and act on the first action}
$P_1$          # "I do not see the apple on the black table. The penguin or the mustard can
             #  hide it. The penguin is larger, so the apple is most likely behind it."
             at[apple]: Lang = "behind the penguin",  at[penguin]: RGBD = x$_T$ (observed from q$_1$),
             movable[penguin] = True,  holding = None,  atRob = q$_1$
(*@\steprule@*)$\vec{a}_1$          pick[penguin, "a grasp that lets me set it down upright"]
             -> place[penguin, table, "far from where the apple may be"]
             -> detect[apple, "facing the black table"]
             -> pick[apple, "a grasp that lets me drop it into the bowl"]
             -> detect[bowl, "toward the wooden desk"]
             -> place[apple, bowl, "inside the bowl"]
(*@\steprule@*)$\Phi$[:1]       Grasp(penguin, x$_T$, g$_T$),  Kin(penguin, x$_T$, g$_T$, q$_2$),  Motion(q$_1$, $\tau_2$, q$_2$)
$\Phi$[1:]       Stable(penguin, table, x$_S$, x$'_T$),  Kin(penguin, g$_T$, x$'_T$, q$_3$),  ...
(*@\steprule@*)$G_1$          grasps = graspgenx.generate(x$_T$)                          # certifies Grasp
             g$_T$: Grasp = agent.choose(grasps, "a grasp that lets me set it down upright")
             q$_2$: Conf = curobo.ik(x$_T$, g$_T$)                             # certifies Kin
             $\tau_2$: Traj = curobo.plan_motion(q$_1$, q$_2$)                 # certifies Motion
             x$'_T$: Lang = "far from where the apple may be"
             g$_A$: Lang = "a grasp that lets me drop it into the bowl"
             x$'_A$: Lang = "inside the bowl"
(*@\steprule@*)$\beta_1$          {g$_T$, q$_2$, $\tau_2$} accepted
(*@\steprule@*)$a_1$          pick[penguin, "a grasp that lets me set it down upright"](x$_T$, g$_T$, q$_2$, $\tau_2$)
             -> holding = penguin,  at[penguin] = g$_T$,  atRob = q$_2$
\end{stepbox}

\textbf{Compiling the whole plan, binding the first action.} The planner compiles the whole plan into $\Phi(\vec{a}_1)$, so the agent sees how its choices are coupled across steps: the penguin's grasp $g_T$ also determines whether it can be set down at $x'_T$, the agent chooses $x'_T$ so that the apple stays reachable, a judgment no factor encodes, and the grasp $g_A$ in turn constrains the placement $x'_A$ in the bowl through \pddl{Kin}. Only the first action is bound with tools. The later variables depend on observations the robot has not made yet, such as the apple's pose once the penguin is gone, so the agent keeps them as language hypotheses and uses them to choose among the candidates of the first action, here a grasp of the penguin that allows a stable placement. The planner accepts $\beta_1$ only because every constraint of \pddl{pick}$[\text{penguin}]$ holds on values a tool has certified.

\begin{stepbox}{typed}{Step 2: the hypothesis is confirmed}
$P_2$          holding = penguin,  atRob = q$_2$,  at[apple]: Lang = "behind the penguin"
(*@\steprule@*)$\vec{a}_2$          place[penguin, table, "far from where the apple may be"]
             -> detect[apple, "facing the black table"]
             -> pick[apple, "a grasp that lets me drop it into the bowl"] -> ...
(*@\steprule@*)$\Phi$[:1]       Stable(penguin, table, x$_S$, x$'_T$),  Kin(penguin, g$_T$, x$'_T$, q$_3$),  Motion(q$_2$, $\tau_3$, q$_3$)
$\Phi$[1:]       constraints of pick[apple] and place[apple, bowl]
(*@\steprule@*)$G_2$          cands = placement(table, "far from where the apple may be")    # 19 placements
             x$'_T$: RGBD = agent.choose(cands);  q$_3$ = curobo.ik;  $\tau_3$ = curobo.plan_motion
(*@\steprule@*)$\beta_2$          {x$'_T$, q$_3$, $\tau_3$} accepted
(*@\steprule@*)$a_2$          place[penguin, table, "far from where the apple may be"](g$_T$, x$'_T$, x$_S$, q$_3$, $\tau_3$)
             -> holding = None,  at[penguin] = x$'_T$,  On[penguin, table] = True,  atRob = q$_3$
             after execution:  sam3("apple") -> 1136 px mask, 693 depth px
             at[apple]: RGBD = x$_A$                                        # language -> certified
\end{stepbox}

\textbf{How a belief changes form.} The apple's location has now passed through two forms of \autoref{fig:variable_ladder}: a language belief, revised between $P_0$ and $P_1$, and a tool-certified RGB-D estimate after the penguin was moved. The planner reads only the form, and the form decides which actions can be grounded: \pddl{pick} requires an RGB-D estimate, so it could not be accepted before this observation. Values tied to the robot's base pose must be recertified after the base moves. This one holds because the robot observed from where it will pick.

\begin{stepbox}{typed}{Step 3: replan from the current belief}
$P_3$          at[apple]: RGBD = x$_A$,  holding = None,  atRob = q$_3$
(*@\steprule@*)$\vec{a}_3$          pick[apple, "a grasp that lets me drop it into the bowl"]  # detect[apple] dropped
             -> detect[bowl, "toward the wooden desk"] -> place[apple, bowl, "inside the bowl"]
(*@\steprule@*)$\Phi$[:1]       Grasp(apple, x$_A$, g$_A$),  Kin(apple, x$_A$, g$_A$, q$_4$),  Motion(q$_3$, $\tau_4$, q$_4$)
$\Phi$[1:]       constraints of detect[bowl] and place[apple, bowl]
(*@\steprule@*)$G_3$          grasps = graspgenx.generate(x$_A$)                          # certifies Grasp
             g$_A$: Grasp = agent.choose(grasps, "a grasp that lets me drop it into the bowl")
             q$_4$: Conf = curobo.ik(x$_A$, g$_A$);  $\tau_4$: Traj = curobo.plan_motion
(*@\steprule@*)$\beta_3$          {g$_A$, q$_4$, $\tau_4$} accepted
(*@\steprule@*)$a_3$          pick[apple, "a grasp that lets me drop it into the bowl"](x$_A$, g$_A$, q$_4$, $\tau_4$)
             -> holding = apple,  at[apple] = g$_A$,  atRob = q$_4$
\end{stepbox}

\textbf{Replanning at every step.} The agent rewrites the problem and plan at every step, so actions whose purpose is already met disappear: \pddl{detect}$[\text{apple}]$ was planned to obtain an RGB-D estimate, which the observation after step~2 already provides. The same mechanism absorbs execution failures. In the run, two base motions toward the apple were aborted by the base controller. The agent observed that the estimate still held from the current pose and picked the apple without moving.

\begin{stepbox}{typed}{Step 4: observe while holding}
$P_4$          holding = apple,  at[bowl]: Lang = "on the wooden desk",  atRob = q$_4$
(*@\steprule@*)$\vec{a}_4$          detect[bowl, "toward the wooden desk"] -> place[apple, bowl, "inside the bowl"]
(*@\steprule@*)$\Phi$[:1]       Motion(q$_4$, $\tau_5$, q$_5$),  Obs(bowl, q$_5$, x$_B$)
(*@\steprule@*)$G_4$          q$_5$: Conf = agent.choose_viewpoint("toward the wooden desk")
             $\tau_5$: Traj = base_planner.plan(q$_4$, q$_5$)                  # certifies Motion
(*@\steprule@*)$\beta_4$          {q$_5$, $\tau_5$} accepted
(*@\steprule@*)$a_4$          detect[bowl, "toward the wooden desk"](b$_B$, x$_B$, q$_5$, $\tau_5$)
             sam3("bowl"): no mask, then 5725 to 7184 px masks
             -> atRob = q$_5$,  at[bowl]: RGBD = x$_B$
\end{stepbox}

\begin{stepbox}{typed}{Step 5: the placement cannot be certified}
$P_5$          holding = apple,  at[bowl]: RGBD = x$_B$,  atRob = q$_5$
(*@\steprule@*)$\vec{a}_5$          place[apple, bowl, "inside the bowl"]
(*@\steprule@*)$\Phi$[:1]       Stable(apple, bowl, x$_B$, x$'_A$),  Kin(apple, g$_A$, x$'_A$, q$_6$),  Motion(q$_5$, $\tau_6$, q$_6$)
(*@\steprule@*)$G_5$          placement(bowl, "inside the bowl") -> no placement reachable with g$_A$ (four attempts)
             FAILURE  Kin of place[apple, bowl]: EMPTY
(*@\steprule@*)$\beta_5$          not accepted;  no action executed
\end{stepbox}

\textbf{The kinematic gap.} The robot knows where the bowl is, but no placement inside it is reachable with the current grasp from where the robot stands. The gap depends on the geometry of the arm, the grasp, and the bowl, so the tools find it, and the planner reports it before the robot moves its arm. TAMPAgent closes it by escalating as in \autoref{subsec:policy}: the agent first revises $G_t$ with other placement candidates, then extends the plan with actions that change the geometry, here a \pddl{detect} that moves the base closer to the bowl. The failure message names the factor, and the agent chooses the fix:
\begin{lstlisting}[style=rec]
{ type: FAILURE,  from: placement,
  text: "No placement inside the bowl is reachable from the current base pose.",
  data: { factor: "Kin(apple, g_A, x_A', q_6) of place[apple, bowl]",  kind: EMPTY } }
\end{lstlisting}

\begin{stepbox}{typed}{Steps 6 and 7: change the geometry and reach the goal}
$P_6$          holding = apple,  at[bowl]: RGBD = x$_B$,  atRob = q$_5$
(*@\steprule@*)$\vec{a}_6$          detect[bowl, "closer to the bowl"]                              # plan extended
             -> place[apple, bowl, "inside the bowl"]
(*@\steprule@*)$\Phi$[:1]       Motion(q$_5$, $\tau_7$, q$_7$),  Obs(bowl, q$_7$, x$_B$)
(*@\steprule@*)$G_6$          q$_7$: Conf = agent.choose_viewpoint("closer to the bowl")
             $\tau_7$: Traj = base_planner.plan(q$_5$, q$_7$)                  # certifies Motion
(*@\steprule@*)$\beta_6$          {q$_7$, $\tau_7$} accepted
(*@\steprule@*)$a_6$          detect[bowl, "closer to the bowl"](x$_B$, x$_B$, q$_7$, $\tau_7$)
             -> atRob = q$_7$,  at[bowl]: RGBD = x$_B$ (recertified)
(*@\steprule@*)$P_7$          holding = apple,  at[bowl]: RGBD = x$_B$,  atRob = q$_7$
(*@\steprule@*)$\vec{a}_7$          place[apple, bowl, "inside the bowl"]
(*@\steprule@*)$\Phi$[:1]       Stable(apple, bowl, x$_B$, x$'_A$),  Kin(apple, g$_A$, x$'_A$, q$_8$),  Motion(q$_7$, $\tau_8$, q$_8$)
(*@\steprule@*)$G_7$          cands = placement(bowl, "inside the bowl")                    # 15 placements
             x$'_A$: RGBD = agent.choose(cands, "drop into the center");  curobo.ik, plan_motion
(*@\steprule@*)$\beta_7$          {x$'_A$, q$_8$, $\tau_8$} accepted
(*@\steprule@*)$a_7$          place[apple, bowl, "inside the bowl"](g$_A$, x$'_A$, x$_B$, q$_8$, $\tau_8$)
             -> holding = None,  at[apple] = x$'_A$,  In[apple, bowl] = True,  atRob = q$_8$
\end{stepbox}

\textbf{Summary.} \autoref{tab:run-summary} lists the run. Both gaps were found by a failed certification before any contact, and each was closed by a change of plan: the epistemic gap by a hypothesis that added actions to reveal the apple, and the kinematic gap by an action that changed the robot's base pose.

\begin{table}[h]
\centering
\small
\begin{tabular}{@{}cllll@{}}
\toprule
Step & First action & Outcome & Gap & Agent's response \\
\midrule
0 & \pddl{detect}[apple] & \texttt{OCCLUDED} & epistemic & hypothesis: behind the penguin \\
1 & \pddl{pick}[penguin] & accepted & & \\
2 & \pddl{place}[penguin, table] & accepted & & apple observed \\
3 & \pddl{pick}[apple] & accepted & & \pddl{detect}[apple] dropped \\
4 & \pddl{detect}[bowl] & accepted & & \\
5 & \pddl{place}[apple, bowl] & \texttt{EMPTY} & kinematic & extend plan: move closer \\
6, 7 & \pddl{detect}[bowl], \pddl{place}[apple, bowl] & accepted & & goal holds \\
\bottomrule
\end{tabular}
\caption{\textbf{The run of \autoref{fig:execution_loop}, step by step.}}
\label{tab:run-summary}
\end{table}

\section{Simulation Tasks}
\label{app:tasks}

\begin{table}[h]
\centering
\small
\caption{\textbf{Kitchen tasks.} We list where the apple is, what the agent must handle before it can pick the apple or place it, and the category of the task (MM: mobile manipulation).}
\label{tab:sim-tasks}
\begin{tabularx}{\linewidth}{@{}l l X c@{}}
\toprule
Task & Apple & What the agent must handle & Category \\
\midrule
T1 & dining table & Nothing. The apple lies next to the plate. & MM \\
T2 & counter & Carry the apple from the counter to the dining table. & MM \\
T3 & refrigerator shelf & Open the closed refrigerator doors before the apple is visible. & MM + occlusion \\
T4 & counter & A cup stands directly in front of the apple, 10.2\,cm between their centers. & MM + non-prehensile \\
T5 & counter & An open drawer below the counter blocks the approach to the apple. & MM + non-prehensile \\
T6 & counter & Both the cup of T4 and the open drawer of T5. & MM + non-prehensile \\
T7 & counter & A cracker box hides the apple from the room side. & MM + occlusion \\
T8 & dining table & A cracker box stands in front of the plate. & MM + non-prehensile \\
T9 & refrigerator shelf & A cup on the shelf stands in front of the apple, behind the closed doors. & MM + non-prehensile + occlusion \\
T10 & end of counter & A mustard bottle hides the apple, and the cracker box of T7 also stands on the counter. & MM + non-prehensile \\
\bottomrule
\end{tabularx}
\end{table}

\begin{table}[h]
\centering
\small
\caption{\textbf{Goal prompt per task.} CaP, GaP, and TAMPAgent receive the same prompt, and nothing else, with no description of the scene.}
\label{tab:task-prompts}
\begin{tabular}{@{}ll@{}}
\toprule
Task & Goal prompt \\
\midrule
T1 to T6, T8, T9 & ``place apple on plate on dining table'' \\
T7, T10 & ``place the apple from counter to plate on dining table'' \\
T11, T13 & ``place bowl from living room to kitchen table'' \\
T12 & ``throw away the used sponge'' \\
\bottomrule
\end{tabular}
\end{table}

\textbf{Scene.} The tasks T1 to T10 share one kitchen with a counter, a stove, a sink, a refrigerator, and a dining table that holds the plate (\autoref{fig:sim-tasks}). The scenes are MuJoCo~\citep{todorov2012mujoco} models of the RBY1 with its XHAND hands and all 50 actuators. The robot always starts at the same pose, facing away from the kitchen, so its first step is to look around. The agent receives only the goal prompt of \autoref{tab:task-prompts} and no description of the scene. The tasks differ in where the apple is and in what stands in its way. T11 to T13 take place in a three-room home with a living room, a kitchen, and a washroom (\autoref{fig:multiroom-tasks}), and the robot starts in the kitchen.

\textbf{Tasks.} \autoref{tab:sim-tasks} lists the kitchen tasks in four categories. The mobile manipulation tasks T1 and T2 require search, pick, carry, and place. The non-prehensile tasks T4 to T6, T8, and T10 add an object or a drawer that blocks the pick or the place, which the agent must move or close first. The occlusion tasks T3 and T7 hide the apple, so the agent must first hypothesize where it is and uncover it. T9 combines both: the apple is inside the fridge behind a cup that the robot must move. We evaluate the tasks on the RBY1, T9 by learning from attempts (\autoref{fig:t9-learning}), and T1 to T6 on the G1 (\autoref{app:g1}). The cracker box is the real-size YCB box~\citep{calli2015ycb} ($7.2 \times 16.4 \times 21.3$\,cm, 0.35\,kg). The mustard bottle is the YCB bottle scaled up by 9.7\%, the smallest scale at which it hides the apple completely in the view from the approach pose in front of the counter. Occlusion depends on the viewpoint, so the apple can still be partly visible from other poses.

\textbf{Success and failure.} An episode succeeds when the planner confirms the goal \pddl{On(apple, plate)} from observed evidence. In T11 and T13, the goal is \pddl{On(bowl,~kitchen\_table)}, and in T12, it is \pddl{In(used\_sponge,~trash\_bin)}. Contact between the robot and the scene does not end an episode. It fails when an object falls or the scene is destroyed, or when its budget of simulated time runs out. The budget is 15 minutes plus 5 minutes for each blocker, and 25 minutes for T3, T7, T9, and T10. The budget counts simulated seconds, so an episode lasts longer in wall-clock time when the simulator runs slower than real time. The episode harness watches the tilt and height of the robot's torso. When the robot falls, the harness stops it, ends the attempt, and restores a fresh scene before the next episode.

\clearpage
\section{Subtask Completion on the RBY1}
\label{app:rby1-subtasks}

\begin{figure}[h]
\centering
\includegraphics[width=\linewidth]{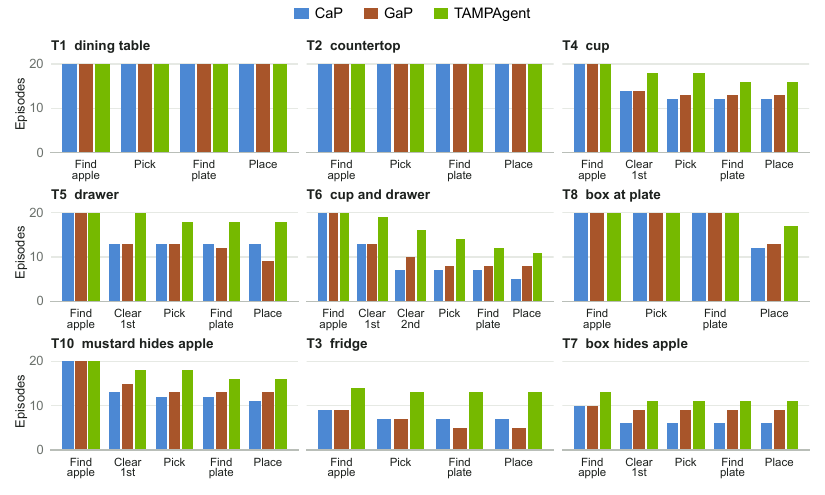}
\caption{\textbf{Subtask completion on the RBY1}, out of 20 episodes per task and method, with the tasks in category order. All three methods find the apple equally often when it is in the open. TAMPAgent separates from CaP and GaP at finding the apple in the fridge (T3) and behind the box (T7), at clearing each blocker (T4 to T6, T10), and at placing next to the box on the dining table (T8), and the gap persists through the remaining subtasks.}
\label{fig:rby1-subtasks}
\end{figure}

\autoref{fig:rby1-subtasks} breaks the success rates of \autoref{fig:rby1-success} down by subtask. The partial success in \autoref{fig:rby1-success} sums, over the 20 episodes, the fraction of applicable subtasks each episode completes.

\clearpage

\section{Additional Results on G1}
\label{app:g1}

We also evaluate TAMPAgent on a G1 humanoid in MuJoCo simulation~\citep{todorov2012mujoco}, on the tasks T1 to T6 of \autoref{sec:benchmark-results} with five episodes per task and method. The G1 is controlled by the whole-body controller SONIC~\citep{luo2025sonic}, which converts base velocity commands and upper-body joint waypoints into loco-manipulation motion. In addition to the tools of \autoref{subsec:formulation}, the agent can call LocateAnything~\citep{wang2026locateanything} for metric 3D localization, nvblox~\citep{millane2024nvblox} for local occupancy mapping, and visual-inertial and LiDAR-inertial odometry~\citep{korovko2025cuvslam,campos2021orbslam3,xu2022fastlio2}. The agent is Codex~\citep{bolin2026codexagentloop} running GPT-5.6 Sol~\citep{openai2026gpt56sol} with medium reasoning effort (the RBY1 experiments use GPT-6-Astra~\citep{openai2026gpt6astra}, at low reasoning effort for all methods). Episodes time out after 15 minutes of simulated time plus 5 minutes for each blocker, and 25 minutes for the occlusion task T3.

\begin{figure}[h]
\centering
\includegraphics[width=\linewidth]{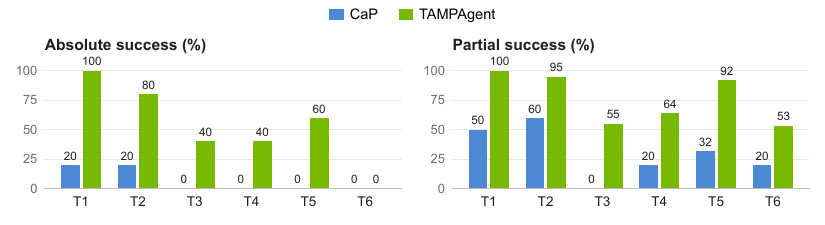}
\caption{\textbf{Success on the G1}, over five episodes per task and method. Partial success is the fraction of applicable subtasks completed, shown per subtask in \autoref{fig:g1-stages}.}
\label{fig:g1-success}
\end{figure}

\begin{figure}[h]
\centering
\includegraphics[width=\linewidth]{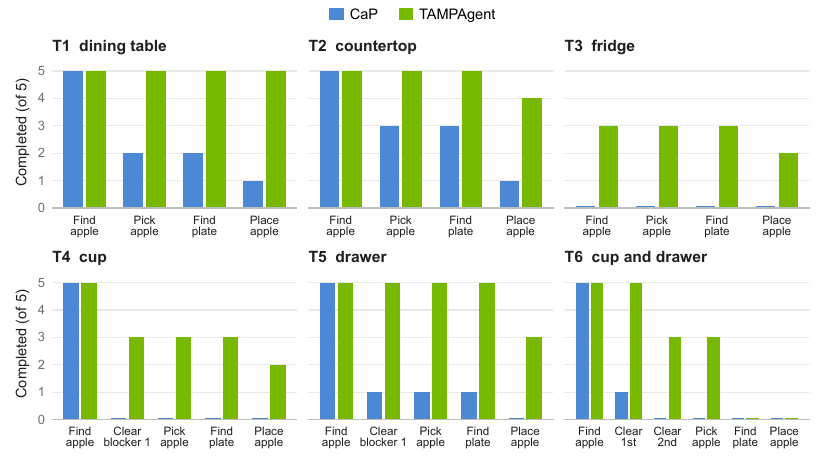}
\caption{\textbf{Subtask completion on the G1}, out of five episodes per task and method. A hairline marks a subtask that applies but was never completed.}
\label{fig:g1-stages}
\end{figure}

The results agree with those on the RBY1. TAMPAgent raises absolute success from 6.7\% to 53.3\% and partial success from 30.3\% to 76.6\% averaged over the six tasks (\autoref{fig:g1-success}). The gap is larger than on the RBY1: on the G1, CaP completes a placement only on T1 and T2, and on the tasks with blockers it finds the apple but rarely clears what stands in front of it (\autoref{fig:g1-stages}).

\section{Hardware and Real-World Setup}
\label{app:hardware}

\begin{figure}[h]
\centering
\includegraphics[width=\linewidth]{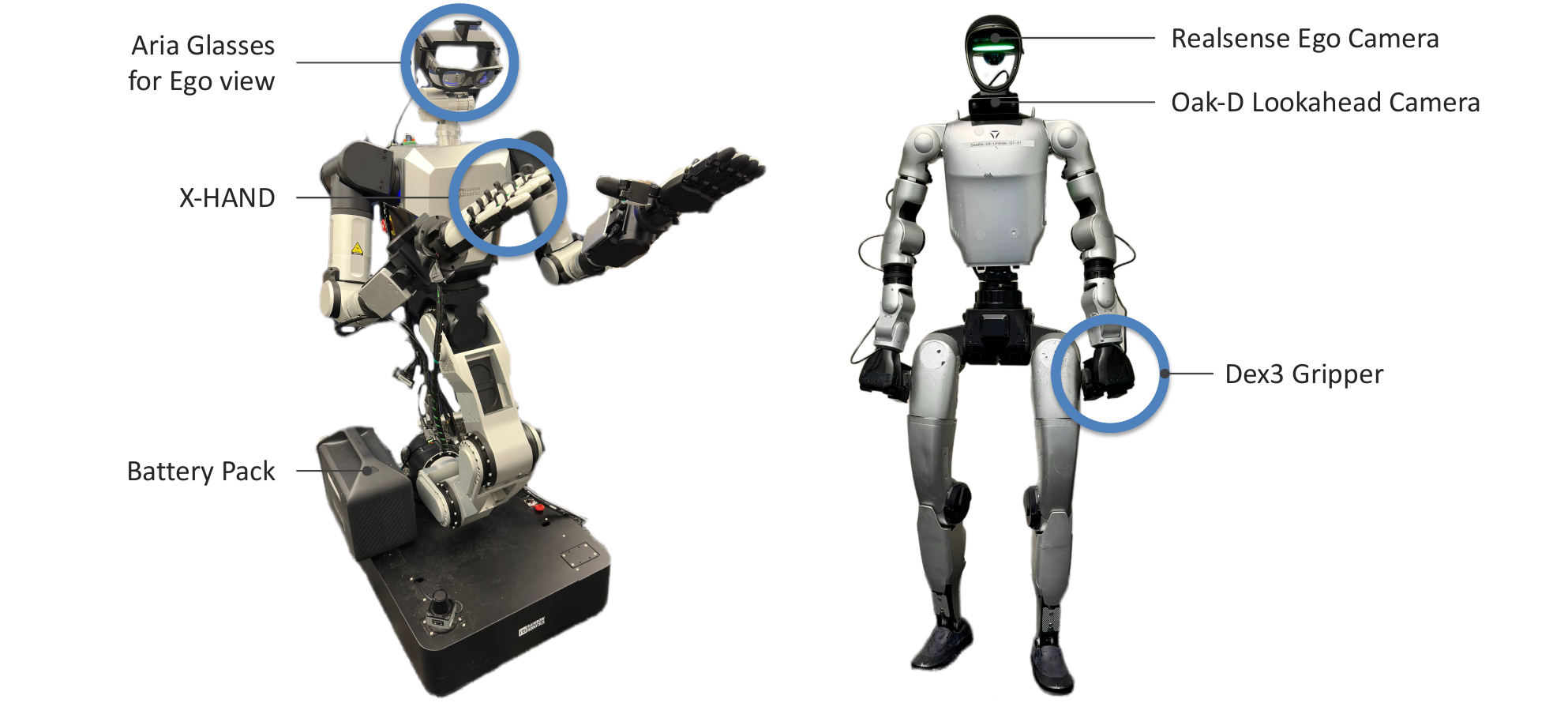}
\caption{\textbf{Robots and sensors.} Left: the Rainbow Robotics RB-Y1 with Project Aria glasses for the ego view, an XHAND dexterous hand, and its battery pack. Right: the Unitree G1 with an Intel RealSense ego camera, an OAK-D look-ahead camera, and Dex3 hands.}
\label{fig:hardware}
\end{figure}

\textbf{RB-Y1.} Except for the G1 run of \autoref{fig:g1-real}, the real-world runs use a Rainbow Robotics RB-Y1~\citep{rainbow2026rby1} (\autoref{fig:hardware}, left), a wheeled bimanual mobile manipulator. One arm ends in an XHAND dexterous hand~\citep{robotera2026xhand}. For the ego view, the robot wears Project Aria glasses~\citep{meta2026aria} on its head, whose stereo images give RGB-D estimates through Fast-FoundationStereo~\citep{wen2026fast}. A look-ahead camera observes from afar. The base is driven by a base motion planner and the arms by cuRobo~\citep{sundaralingam2026curobov2}.

\textbf{Unitree G1.} The G1 experiments use a Unitree G1 humanoid~\citep{unitree2026g1} (\autoref{fig:hardware}, right) with Dex3 hands. An Intel RealSense D435i~\citep{intel2026d435i} serves as the ego camera and a Luxonis OAK-D Pro~\citep{luxonis2026oakdpro} as the look-ahead camera. The whole-body controller SONIC~\citep{luo2025sonic} turns base velocity commands and upper-body joint waypoints into loco-manipulation motion.

\textbf{Compute.} The agent runs in the cloud as Codex~\citep{bolin2026codexagentloop} with GPT-6-Astra~\citep{openai2026gpt6astra}, in a sandbox that can reach only the planner and the tools. The planner and the tools run on a local workstation connected to the robot.  Every episode is recorded with its events, tool outputs, factor graphs, and camera streams, which is the source of the trace in \autoref{app:trace}.

\textbf{Scenes.} The RB-Y1 scenes are in a lab setup. A black table holds the objects to find, among them an apple, a mustard bottle, and a toy penguin, any of which can hide another from the robot's starting view. A wooden desk elsewhere in the room holds the receptacles, a wooden bowl and a blue basket. The robot starts every episode away from the objects, so each task begins with exploration, and no scene description is given to the agent. The tasks are instructions of the form \emph{``place the apple from the black table in the wooden bowl''} or \emph{``place the mustard bottle in the blue basket''}, with blockers arranged so that the target is hidden or cannot be grasped directly. In the G1 run, the goal is to hold a yellow can whose grasp a toy dog blocks (\autoref{fig:g1-real}).

\section{System Architecture}
\label{app:architecture}

\begin{figure}[h]
\centering
\includegraphics[width=\linewidth]{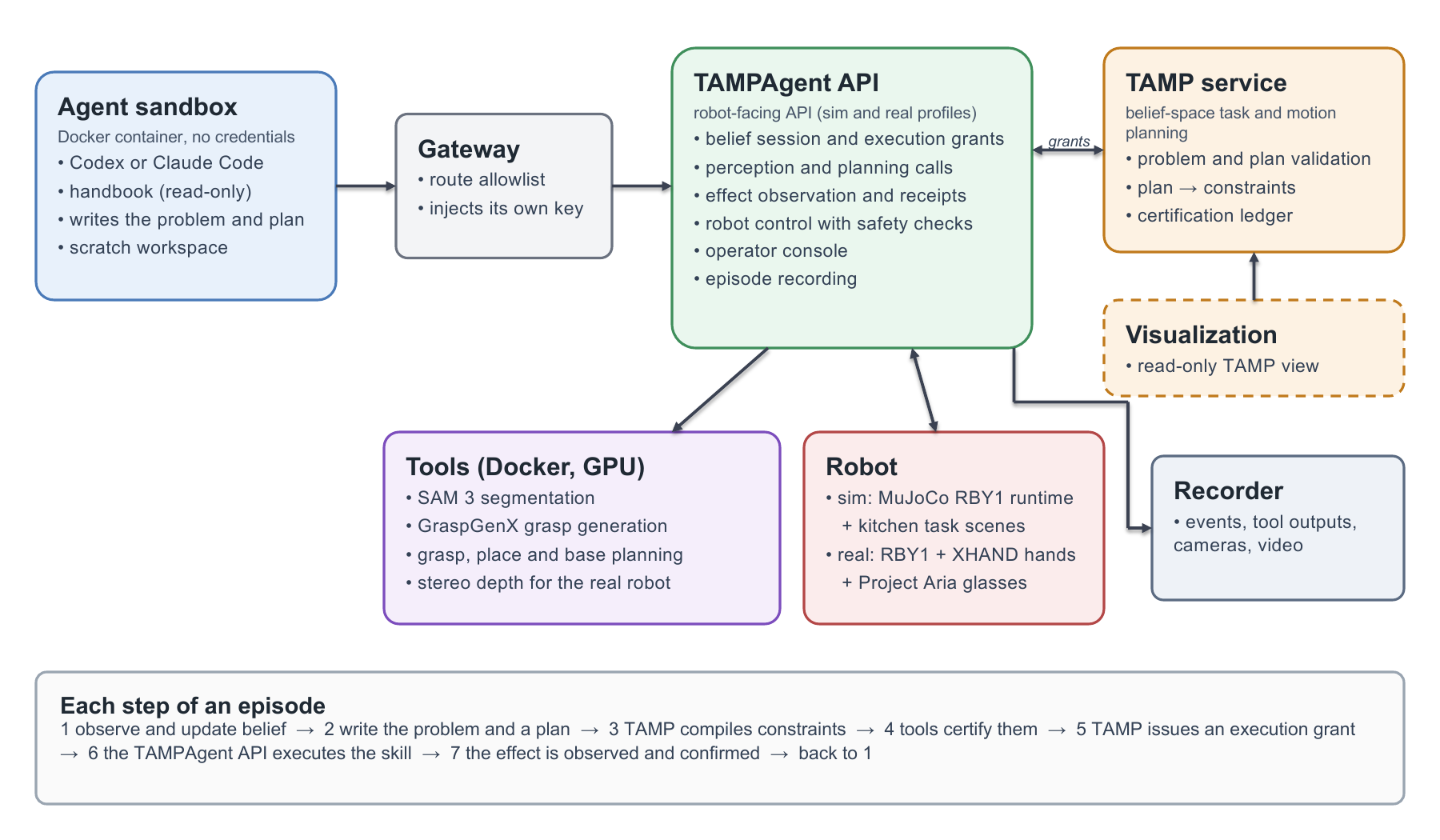}
\caption{\textbf{System architecture.} We run the agent in a sandbox that reaches the system only through a gateway. The TAMPAgent API is the robot-facing API. It holds the belief session, forwards perception and planning calls to the tools, executes skills on the robot only under grants from the TAMP service, and records the episode. The visualization reads the TAMP session without changing it.}
\label{fig:architecture}
\end{figure}

\textbf{Components.} TAMPAgent runs as a set of local services and one sandboxed agent (\autoref{fig:architecture}). The services are (1) the \emph{gateway}, the agent's only route into the system (\autoref{app:sandbox}), (2) the \emph{TAMPAgent API}, which faces the robot, (3) the \emph{TAMP service}, the belief-space planner of \autoref{subsec:formulation}, (4) the \emph{tools}, GPU services for perception and motion planning, (5) the \emph{robot}, either the MuJoCo runtime of \autoref{app:tasks} or the RBY1 hardware of \autoref{app:hardware}, and (6) the \emph{recorder} for every event, tool output, and camera stream of an episode. The TAMPAgent API checks each command against the TAMP session, so manipulation and base motion run only under a TAMP grant and within its window. The operator can stop the robot from a console at any time.

\textbf{One step.} Each step of an episode passes through the services in a fixed order. (1) The agent observes through the TAMPAgent API and writes its belief into the TAMP session, that is, the inventory of entities, the location hypotheses, and the observed facts. (2) It proposes a plan over the planner's operators (\pddl{explore}, \pddl{approach}, \pddl{pick}, \pddl{place}, \pddl{move\_holding}, \pddl{reorient\_holding}, \pddl{open\_fixture}, \pddl{close\_fixture}, and the \pddl{confirm} checks) and locks it. (3) The TAMP service compiles the next step of the plan into constraints. (4) The agent calls the tools through the TAMPAgent API and submits their outputs as evidence for each constraint. (5) Once every constraint of the step is certified, the TAMP service issues an execution grant, and the TAMPAgent API executes the granted skill. (6) The TAMPAgent API observes the effect. The TAMP service adds a confirmed effect to the belief. An unconfirmed effect is settled, and the agent replans.

\begin{table}[h]
\centering
\small
\caption{\textbf{Tools.} We list each tool service, the model or method behind it, and what its output certifies. The planners sign their outputs, and the TAMPAgent API executes only trajectories with a valid signature.}
\label{tab:tools}
\begin{tabularx}{\linewidth}{@{}l l X@{}}
\toprule
Tool & Model or method & Output \\
\midrule
Segmentation & SAM 3~\citep{carion2025sam3segmentconcepts} & masks for a text prompt and the observed geometry of the object \\
Grasp generation & GraspGenX~\citep{han2026graspgenx} & grasp candidates from the observed geometry \\
Grasp planning & cuRobo~\citep{sundaralingam2026curobov2} & signed, collision-checked trajectories to a grasp \\
Place planning & cuRobo~\citep{sundaralingam2026curobov2} & placement candidates and signed trajectories for the held object \\
Base planning & RGB-D route planner & signed base routes and bounded motion windows \\
Stereo depth (real) & Fast-FoundationStereo~\citep{wen2026fast} & depth from the rectified stereo pair of the Aria glasses \\
\bottomrule
\end{tabularx}
\end{table}

\textbf{Tools.} \autoref{tab:tools} lists the tool services. The tools return perception results or plans and never move the robot. A plan reaches the robot only as part of a granted skill call, and the TAMPAgent API checks the plan's signature before the robot executes the trajectory.

\begin{table}[h]
\centering
\small
\caption{\textbf{Simulation and real-robot profiles.} We list where the two profiles differ. They share the planner, the gateway, the sandbox, the recorder, and the agent's documentation apart from the pages named here.}
\label{tab:profiles}
\begin{tabularx}{\linewidth}{@{}l X X@{}}
\toprule
 & Simulation & Real robot \\
\midrule
Robot & MuJoCo RBY1 runtime with the task scenes & RBY1 with an XHAND hand and Aria glasses \\
Perception & SAM 3, simulated depth & SAM 3 and Fast-FoundationStereo depth \\
Holding after a pick & observed by the runtime & asserted by the agent from its observations \\
Doors and drawers & fixture skills available & not available \\
Base motion & planned routes and straight legs up to 0.5\,m & straight legs up to 1.5\,m, no route planner \\
Torso height & about 1.00 to 1.29\,m, with empty hands only & about 1.02 to 1.31\,m \\
Episode verdict & from the planner's goal check or the time budget & from the operator \\
\bottomrule
\end{tabularx}
\end{table}

\textbf{Profiles.} The same code runs in simulation and on the real robot, and the two profiles can run at the same time on one machine with a shared tool stack. \autoref{tab:profiles} lists where they differ. The largest difference is how the planner learns that the robot holds an object. In simulation, the runtime measures it. On the real robot, the agent asserts it from its observations after a pick, and the assertion enters the belief as the agent's claim.

\textbf{Credentials.} Every service authenticates to the others with keys and tokens generated once at setup and stored in files readable only by their owner. The agent holds none of them. The gateway removes any credential the agent supplies and attaches its own fixed agent identity (\autoref{app:sandbox}).

\section{Agent Sandbox}
\label{app:sandbox}

\textbf{Isolation.} The agent is Codex~\citep{bolin2026codexagentloop} running GPT-6-Astra~\citep{openai2026gpt6astra} at low reasoning effort, as do the CaP and GaP agents (\autoref{app:baselines}), inside a Docker container with all Linux capabilities dropped and no privilege escalation. The container has no Docker socket and no mount of host state. The documentation, skill pages, and goal prompt are mounted read-only, and the agent writes only to a scratch directory. The container cannot reach the robot, the cameras, the simulator, or any planner backend directly. Host firewall rules for the container's network (1) admit only the gateway's port on the host, (2) reject all traffic to private and link-local address ranges, and (3) leave public access for model inference. IPv6 is disabled on that network. The firewall rules are installed by a short-lived helper container, so the agent's container never holds the right to change them.

\textbf{Gateway.} The gateway is the only local service the agent can reach. It authorizes the method, the path, and the body of every request before it opens a connection to the TAMPAgent API. It accepts only routes on a positive allowlist of 87 routes in simulation and 77 on the real robot, taken from the same registry that generates the agent's documentation, so the documentation and the allowlist cannot drift apart. It removes any caller-supplied credentials or backend selection and forwards the request under a fixed agent identity. It forwards one request per connection, bounds request bodies at 1\,MiB, refuses streamed and ambiguous requests, and never retries a request. Every JSON reply passes through a projection that removes host paths and ground-truth object identities, and binary replies must be JPEG or PNG images. On the real robot, outside an active evaluation, the gateway blocks every call that changes state except stopping the robot. The agent cannot start or end sessions, record episodes, declare success or failure, or see other episodes.

\textbf{Episodes.} An episode harness runs each episode in six steps. (1) It checks the configuration, the agent container, and that no other episode is recording. (2) It archives and clears the previous agent and launches a fresh one with an empty workspace and no memory of earlier episodes. (3) It prepares the scene. In simulation, the runtime restarts on the task's scene. On the real robot, the harness waits for the operator to prepare it. (4) It starts the belief session and the recording, and it hands the task to the agent as a goal. (5) It waits until the goal is reached, the time budget runs out, or the operator ends the episode. (6) It stops the robot, checks that the robot is still, closes the session, and archives the agent's workspace and token usage. Nobody messages the agent during an episode, so each episode measures what the agent finds on its own.

\textbf{Documentation.} The agent sees the following files. It is told to read the run book and the page for its current stage before acting. Both profiles share one handbook, and a profile replaces a page or inserts a short part only where the robots differ.
\begin{lstlisting}[style=trace]
/workspace
  AGENTS.md, CLAUDE.md   entry page: read the handbook before using robot tools
  handbook/              read-only, built per profile (simulation or real robot)
    AGENT.md             contract: role, the loop, workspace and script format, conduct
    AGENT_API.md         every allowlisted route with its category and effect
    PIPELINE.md          each skill: needs, admission tools, dispatch, completion tools
    AGENT_CUSTODY.md     how Holding enters, leaves, and is corrected after a pick or place
    CAPTURE_SELECTION.md choosing among capture candidates; drawing a mask if SAM 3 fails
    tampagent_api.py         client: one request per call, saved to disk, no retries
    commands/goal.md     the goal page
    skills/
      RUNBOOK.md         call order, waiting, replay, outcome -> next action
      AUTHORING.md       program schema: objects, goals, location hypotheses with priors
      EVIDENCE.md        captures, certifying sensed/identity/location, blockers
      PLANNING.md        propose, adopt, and lock plans; candidate beliefs; grants
      MOVE.md            explore, approach, move while holding; bounded base legs
      PICK.md, PLACE.md  reach, close or open, trace back, verify, recover
      FIXTURES.md        doors and drawers (simulation); none on the real robot
      SCHEMAS.md         exact request keys and schemas for every route
  work/                  the agent's own scripts and saved requests and replies
\end{lstlisting}
The handbook contains no strategy and no task examples. A build step checks that every route the pages name is on the profile's allowlist, that the schema page has exactly one entry per allowlisted route, and that no page names an object, surface, or scene from an earlier run.

\textbf{What the agent can write.} The routes fall into seven groups, which match the stages of \autoref{fig:execution_loop}: (1) authoring the problem (validate and populate a program, amend destinations), (2) beliefs and evidence (declare inventory, location classes, and access assessments, certify captured evidence, assert custody), (3) planning (grounded actions, propose plans with a belief over them, extend the problem, adopt, lock, unlock), (4) perception (camera images, captures, target selection, drawn masks, head and wrist views), (5) grasp and placement planning (candidates, confirmation, refinement), (6) execution (grant requests and the bounded skills they authorize), and (7) verification (observe and confirm effects, execution history, goal events). The agent may author (a) the program, that is, objects, supports and containers, goals in \pddl{On}, \pddl{In}, or \pddl{holding}, and location hypotheses with priors, (b) the inventory and access assessments, (c) plans and their extensions, (d) probability beliefs over plans and over tool candidates, which must leave some mass to an unknown outcome, and (e) its own scripts. It may not (a) author sensor facts, feasibility, grants, or effects, (b) move without an accepted grant, or manipulate without a locked plan, (c) change thresholds, (d) call routes outside the allowlist, or (e) repeat an uncertain motion under a new request. Every script it writes must state what it believes, what it intends, and what it achieved:
\begin{lstlisting}[style=trace]
# <what do you currently believe> ...
# <what you want to do> ...
<code>
# <what you achieved> ...
\end{lstlisting}

\textbf{Examples.} An abridged script written by the agent in a real run authors the problem, with two location hypotheses for an unseen object:
\begin{lstlisting}[style=trace]
# Belief: Cup, plate, source table and adjacent brown table are visually observed;
#         pink object remains unobserved.
# Intent: Validate and populate one full task program, declare only observed inventory.
p = dict(task_id='serve_pink_and_cup', objects=[
      dict(object_id='pink_object', identity='unknown', destination='plate', relation='On',
           hypotheses=[hyp('source_table', 0.6), hyp('brown_table', 0.4)]),
      dict(object_id='cup', identity='unknown', destination='brown_table', relation='On',
           hypotheses=[hyp('source_table', 1.0)]), ...],
      goals=[dict(relation='On', args=['pink_object', 'plate']),
             dict(relation='On', args=['cup', 'brown_table'])])
v = m.call('validate', '/perception/belief_tamp/validate_program', {'program': p})
r = m.call('populate', '/perception/belief_tamp/populate_program',
           dict(..., program=p, program_digest=v['program_digest']))
# Achieved: Program validated and populated once; inventory accepted at revision 2.
\end{lstlisting}
When it selects among grasp candidates, the agent attaches its belief over them. The planner accepts the grasp only if the selected candidate was certified by the grasp tool:
\begin{lstlisting}[style=trace]
"agent_candidate_belief": {
  "candidate_priors": [{"candidate_id": "<candidate ID>", "probability": 0.9}],
  "unknown_mass": 0.1, "absent_mass": 0.0, "other_mass": 0.0,
  "first_choice_candidate_id": "<candidate ID>" }
\end{lstlisting}
A motion is requested by naming the locked plan, its step, and the skill endpoint it binds. The reply is a grant that the skill call must carry:
\begin{lstlisting}[style=trace]
{ "schema": "tampagent.request_execution_grant_request.v1",
  "request_id": "<stable ID>", "session_id": "<session>",
  "expected_plan_id": "<plan>", "expected_plan_digest": "<digest>",
  "step_id": "<step>", "operator": "<operator>", "endpoint": "<bound endpoint>", "body": {} }
\end{lstlisting}

\raggedbottom
\section{Baselines: Code as Policies and Graph as Policy}
\label{app:baselines}

\textbf{Choice of baseline.} We compare against Code as Policies (CaP)~\citep{liang2023codeaspolicies}, in which the agent writes programs that call perception and control directly. It is the paradigm behind most current agentic robot systems~\citep{huang2023voxposer,chen2026gap,fu2026capx}, and it is the closest alternative to TAMPAgent because both let the agent decide which tools to call and when the robot moves. We implement CaP on our own stack rather than running the original code, so that the two methods share the robot, the simulator, the tools, the model, and the sandbox. The comparison then isolates the one thing that differs, which is the belief-space planning representation the agent works in.

\textbf{Implementation.} The CaP agent is the same Codex agent running GPT-6-Astra at the same low reasoning effort, in the same container, behind the same gateway, and under the same episode harness (\autoref{app:sandbox}). It receives the task instruction as its goal, starts each episode with an empty workspace, and nobody messages it during the episode. It writes every computation as a Python script in its workspace, runs it, and saves every reply to a file, as TAMPAgent does. Its handbook keeps the pages that describe the robot and its tools and drops the pages that describe the planning model:
\begin{lstlisting}[style=trace]
/workspace/handbook
  AGENT.md             contract: role, the loop, workspace and script format, conduct
  AGENT_API.md         every allowlisted route with its category and effect
  CAPTURE_SELECTION.md choosing among capture candidates
  skills/
    RUNBOOK.md         call order, waiting, replay
    SCHEMAS.md         exact request keys, replies, and refusals for every route
    MOVE.md            looking, direct base legs, planned base routes
    PICK.md, PLACE.md  reach, close or open, trace back, look
    FIXTURES.md        doors and drawers
\end{lstlisting}
The run book prescribes the loop: (1) sweep the head camera over its range and look at each view, (2) capture the target, and (3) move, pick, place, or operate a door or drawer one stage at a time, looking at the camera after each stage. The agent marks the goal complete when fresh images from the robot's camera show the requested result with the hand withdrawn.

\textbf{Tools.} The CaP allowlist contains the camera, state, perception, base motion, manipulation, and fixture routes, with the same backends as TAMPAgent (\autoref{sec:method}). A capture segments a named object in a frozen RGB-D frame and returns its position, range, and whether it lies within the grasp envelope. A grasp plan returns grasp options for the captured object, among which the agent selects one. A placement request samples reachable placements of the held object on a captured destination. A base plan proposes a stance from the obstacles in the observed views. A pick, for instance, is the following sequence of calls, each followed by a look at the camera:
\begin{lstlisting}[style=trace]
/skill/head (sweep)  ->  /perception/capture {prompt: "apple"}
/perception/grasp_plan  ->  /perception/grasp_plan {selected_part_id}
/perception/grasp_execute_replay  ->  /perception/grasp_close  ->  /perception/grasp_trace_back
/camera/color        the apple moved with the hand
\end{lstlisting}
The allowlist omits every route for authoring a problem, declaring beliefs and evidence, proposing, adopting, and locking plans, and requesting execution grants.

\textbf{Differences from TAMPAgent.} The two methods see the same robot through the same tools, and they differ in what the agent writes and in what it is told about each action. \autoref{tab:cap-vs-tampagent} lists the differences.
\par\noindent\begin{minipage}{\linewidth}
\centering
\small
\renewcommand{\arraystretch}{1.2}
\begin{tabularx}{\linewidth}{@{}>{\raggedright\arraybackslash}p{0.2\linewidth}>{\raggedright\arraybackslash}X>{\raggedright\arraybackslash}X@{}}
\toprule
 & \textbf{CaP} & \textbf{TAMPAgent} \\
\midrule
What the agent writes & Scripts that call tools, chosen one step at a time & A problem $P_t$, a plan, and scripts that bind its parameters with tools \\
Unknown locations & Searched one at a time, the next proposed after the current one fails & Ranked language hypotheses over every needed object, written before acting \\
Dependencies between steps & Kept in the agent's context & Explicit in the plan skeleton and its CSP, so a blocker's removal can be planned before any motion \\
Handbook & Robot and tool pages & Adds pages for authoring, evidence, planning, the skill pipeline, and custody \\
\bottomrule
\end{tabularx}
\captionof{table}{\textbf{CaP and TAMPAgent on the same stack.} Both use the same model, sandbox, robot, and tools, both at low reasoning effort.}
\label{tab:cap-vs-tampagent}
\end{minipage}

The representation shapes how the agent reasons. Writing $P_t$ asks the agent to state its beliefs and goals in terms of objects, relations, and hypotheses, and the planner's grounded actions put plans such as moving a blocker among its options from the start. A CaP agent holds the same information only in its own context, so it reasons about each step as it reaches it. We attribute the gaps on the tasks with a hidden apple or a blocker, and TAMPAgent's lower token usage, to this difference (\autoref{sec:benchmark-results}).

\textbf{Graph as Policy.} We port Graph as Policy (GaP)~\citep{chen2026gap} onto the CaP stack, so that the two baselines differ only in how the agent represents its policy. GaP uses the same robot, simulator, routes, and handbook as CaP, with an added page on graphs, and runs the original graph runtime unmodified. The agent delegates graph authoring to two helper agents, an author that writes a graph of tool calls and a reviewer that checks it, both Codex processes with the same model at low reasoning effort. A graph is validated before it sends any request, and outside a graph the agent may only read the robot's state or stop it, so every motion runs through a validated graph. We keep GaP's graph, its validation, the author and reviewer agents, and the execution traces. We drop rehearsal in a separate simulator, checkpoints, and GaP's own skill library, which assume known object models and fixed arms. In the episodes we inspected, the agent had a single generic graph built that observes, runs the action it names, and observes again, and then ran it once per action, up to 179 times in an episode. The structure of the task thus stayed in the agent's conversation rather than in the graph. GaP is more structured than CaP, but its agent must come up with each computation graph entirely by itself. In TAMPAgent, the planner already provides the symbolic information the graph needs, the grounded actions of the plan and, through $\Phi(\vec{a}_t)$, the constraints to certify and the tool that certifies each, so the agent only composes those tools.

\end{document}